\documentclass[10pt,twocolumn,letterpaper]{article}

\usepackage{lineno}
\usepackage[pagenumbers]{iccv} %
\usepackage[normalem]{ulem}
\definecolor{sdf}{RGB}{230, 255, 230}
\definecolor{mvs}{RGB}{230, 240, 255}
\usepackage{amsmath}
\usepackage{amssymb}
\usepackage{graphicx}
\usepackage{booktabs}
\usepackage{float}
\usepackage{overpic}
\usepackage{multirow}
\usepackage{stfloats}
\usepackage{soul}
\usepackage{wrapfig}
\usepackage{pict2e}
\usepackage[ruled,vlined]{algorithm2e}

\definecolor{cvprblue}{rgb}{0.21,0.49,0.74}
\definecolor{iccvblue}{rgb}{0.21,0.49,0.74}

\usepackage{hyperref}
\hypersetup{breaklinks,colorlinks,citecolor=cvprblue}

\def\method{J-NeuS}

\def\paperID{9825} %
\def\confName{ICCV}
\def\confYear{2025}

\title{J-NeuS: Joint field optimization for Neural Surface reconstruction in urban scenes with limited image overlap}

\author{   
    Fusang Wang$^
    {1,2}$ \qquad
    Hala Djeghim$^
    {1,3}$ \qquad
    Nathan Piasco$^{1}$ \qquad
    Moussab Bennehar$^{1}$ \qquad
    Luis Roldão$^{1}$ \\ 
    Yizhe Wu$^{1}$ \qquad
    Fabien Moutarde$^{2}$  \qquad
    Désiré Sidibé$^{3}$  \qquad
    Dzmitry Tsishkou$^{1}$ \qquad \\
    $^{1}$Noah's Ark, Huawei Paris Research Center, France \\
    $^{2}$CAOR, Mines-Paris PSL, France \qquad $^{3}$IBISC, Evry Paris-Saclay University, Franc
}

\begin{document}

\twocolumn[{%
\renewcommand\twocolumn[1][]{#1}%
\maketitle
\vspace{-1.2cm}

\begin{center}
	\centering
    \begin{overpic}[width=.75\linewidth]{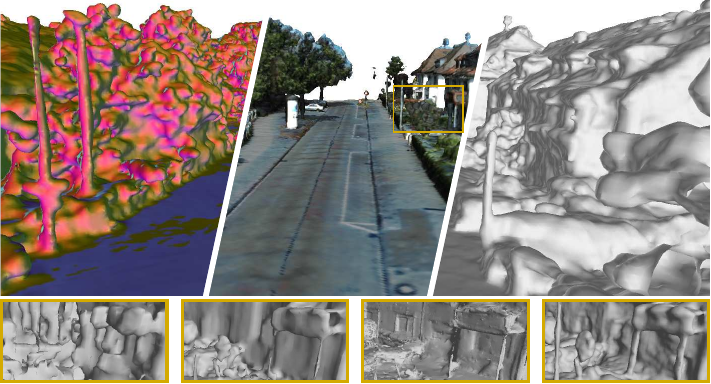}
    \put(5, -3){StreetSurf~\cite{streetsurf}}
    \put(31, -3){ViiNeuS~\cite{scilla}}
    \put(58.5, -3){GoF~\cite{Yu2024GOF}}
    \put(78.5, -3){\method~{(ours)}}
    \end{overpic}
  \vspace{2mm}
	\captionof{figure}{We introduce \textbf{\method}, a novel hybrid implicit surface reconstruction method specifically designed for large-scale driving sequences with limited camera overlap. Extensive experiments on four major driving datasets demonstrate the superiority of \method's mesh (top, form left to right: mesh normals, textured mesh and shaded mesh) over previous state-of-the-art methods (bottom).}
	\label{fig:teaser}
\end{center}

}]

\maketitle
\begin{abstract}
Reconstructing the surrounding surface geometry from recorded driving sequences poses a significant challenge due to the limited image overlap and complex topology of urban environments. SoTA neural implicit surface reconstruction methods often struggle in such setting, either failing due to small vision overlap or exhibiting suboptimal performance in accurately reconstructing both the surface and fine structures.
To address these limitations, we introduce \textbf{\method}, a novel hybrid implicit surface reconstruction method for large driving sequences with outward facing camera poses. \method~leverages cross-representation uncertainty estimation to tackle ambiguous geometry caused by limited observations. Our method performs joint optimization of two radiance fields in addition to guided sampling achieving accurate reconstruction of large areas along with fine structures in complex urban scenarios. Extensive evaluation on major driving datasets demonstrates the superiority of our approach in reconstructing large driving sequences with limited image overlap, outperforming concurrent SoTA methods.

\end{abstract}
   
\section{Introduction}
\label{sec:intro}
Accurate 3D surface reconstruction of large urban scenes is essential for many challenging autonomous driving applications, such as scene relighting~\cite{Lightsim}, sensor simulation~\cite{yang2023unisim}, and 3D object insertion~\cite{fegr}. However, achieving high-quality reconstructions in driving environments remains a significant challenge, where the difficulty primarily stems from two key factors:
\begin{itemize}
    \item \textbf{Complex outdoor geometry}: Urban scenes often feature arbitrary object arrangements, including large textureless planar surfaces and intricate fine structures.
    \item \textbf{Geometric ambiguity from outward facing sensors}: Limited camera overlap and the linear trajectory of vehicles introduce significant uncertainty in the reconstruction process.
\end{itemize}

\noindent Neural Radiance Fields (NeRF)~\cite{2020nerf} have emerged as a powerful approach for 3D scene reconstruction in such settings. Leveraging differentiable volume rendering, NeRF enables accurate reconstruction from images and their associated camera poses, effectively avoiding the error accumulation typically seen in traditional multi-view stereo pipelines.
Although NeRF-based methods have demonstrated impressive results in capturing high frequency scene details ~\cite{mip-nerf-360, mueller2022instant}, its under-constrained optimization problem leads to bad scene geometry, especially when vision cues are limited~\cite{niemeyer2022regnerf, structnerf}. To adapt to large scenes with complex structures, either 3D priors, such as LiDAR pointcloud, or strong assumptions are introduced to constrain the original optimization problem~\cite{rematas2022urban, deng2022depth, wang2022neuris, wang2023planerf}.

Neural implicit surface methods~\cite{NeuS, neus2, VolSDF}, overcome the limitation of NeRF by replacing the volumetric density field with a Signed Distance Function (SDF). The SDF formulation represents the surface at its zero level sets and is regularized using the Eikonal constraint. Current neural SDF methods demonstrate high fidelity reconstruction quality in object-centric scenes with large texture-less surfaces~\cite{li2023neuralangelo, Yu2022MonoSDF}. However prevalent SDF methods like Neuralangelo~\cite{li2023neuralangelo} work better on landmark-centric scenes with large image overlaps but fail to reconstruct urban scenes captured by vehicle-mounted cameras due to limited observation overlap~\cite{streetsurfgs, scilla}. Moreover, SDF-based methods often struggle to preserve fine structural details due to biased depth estimation and over-regularization of geometric constraints~\cite{wang2024neurodin, zhang2023towards}, an issue that is critical for downstream autonomous driving applications.

To achieve high-quality surface reconstruction for autonomous driving -- \textit{capturing both fine structures and large surfaces} -- recent approaches integrate volumetric and SDF representations. These methods partition scenes into distinct regions and apply specialized reconstruction~\cite{streetsurf} or sampling strategies~\cite{turki2024hybridnerf, wang2023adaptive} tailored to each region's unique characteristics. Additionally, other methods~\cite{scilla} demonstrate that SDF representations can benefit from volumetric initialization, enabling faster convergence and improved geometric fidelity. However, these solutions often result in suboptimal geometry quality due to overly strong geometric assumptions about the scene or reliance on coarse initialization that introduce noise from volumetric prediction.

In order to accurately combine the strengths of both volumetric and SDF representation, we propose an uncertainty estimation framework that identifies a noisy predictions arising from geometric ambiguity. Specifically, we estimate two types of uncertainty -- \textit{photometric uncertainty and geometric uncertainty} -- to jointly train the volumetric and SDF models. These uncertainty measures guide the sampling process, ensuring that each representation is deployed where it performs better while mitigating over-regularization in the SDF to facilitate fine structure learning. Our approach effectively handles the complex geometry of urban environments, enabling efficient rendering and precise surface reconstruction. We evaluate our proposed solution on four public driving datasets: KITTI-360~\cite{Kitti}, Pandaset~\cite{pandaset}, Waymo Open Dataset~\cite{Waymo}, and nuScenes~\cite{nuscenes}, demonstrating  robust reconstructions of intricate urban geometry with limited image overlap.

\noindent The main contributions of our method are the following:
\begin{itemize}
    \item \textbf{Joint optimization of NeRF and SDF}: We propose a framework that fuses volumetric (NeRF) and surface (SDF) representations so each excels in regions where it is better.
    
    \item \textbf{Guided Ray Sampling}: We introduce a novel sampling strategy that leverages cross-representation uncertainty to tackle ambiguous geometric cues, enabling faster and more accurate surface reconstruction.
    
    \item \textbf{Adaptive Relaxation on geometry regularization}: We dynamically relax the Eikonal constraint and monocular cues in uncertain regions to avoid over-smoothing, ensuring complete fine-structure reconstruction.
\end{itemize}

\section{Related work}
\paragraph{Neural implicit surface reconstruction}
Traditional surface reconstruction techniques, such as Multi-View Stereo (MVS)~\cite{openMVG}, have long been the cornerstone of 3D reconstruction tasks. However, their multi-step pipelines are prone to error accumulation, often resulting in incomplete or inaccurate reconstructions. In contrast, neural implicit surface reconstruction approaches adapt volume rendering for more accurate surface estimation~\cite{NeuS, VolSDF}, reducing the need for manual post-processing. Subsequent advances target faster training~\cite{neus2} or to and more complex geometry~\cite{li2023neuralangelo, wang2024neurodin}. Notably, Neuralangelo~\cite{li2023neuralangelo} and Neurodin~\cite{wang2024neurodin} achieve highly fidelity reconstruction of complex geometries in object-centric scenes where large observation overlaps are present. However, they face significant challenges when applied to autonomous driving scenarios~\cite{scilla, streetsurfgs}, where the camera trajectory is linear and image overlap is limited.

\paragraph{Urban outdoor surface reconstruction.}

To enhance surface reconstruction in autonomous driving scenes with limited image overlap, recent methods typically rely on 3D supervision~\cite{fegr, urban-radiance-fields}, strong geometric priors~\cite{streetsurf, wang2023planerf}, or monocular supervision~\cite{streetsurf, scilla}. While these approaches effectively model large planar areas, they often struggle to capture fine details due to 
strong geometric assumptions~\cite{wang2023planerf}, inconsistencies in monocular predictions~\cite{streetsurf} or over-regularization on intricate features~\cite{scilla}.

Recent approaches have also explored 3D Gaussian splatting for surface reconstruction~\cite{chen2023neusg, Guedon_2024_CVPR, chen2024pgsrplanarbasedgaussiansplatting}. These methods, however, often rely on heavy post-processing techniques, such as Poisson Surface Reconstruction~\cite{kazhdan2006poisson}, to extract the final surface, adding computational complexity. Gaussian Opacity Field (GoF)~\cite{Yu2024GOF} offers a direct approach for mesh extraction by explicitly learning level sets, avoiding the need for post-processing. However, these methods may suffer from excessive memory usage when reconstructing large-scale driving scenes. To address this, StreetSurfGS~\cite{streetsurfgs} introduces a planar-based octree representation and segmented training, significantly reducing memory usage and making the method more suitable for large-scale scene reconstruction in autonomous driving. Nonetheless, it still requires TSDF fusion and additional mesh cleaning to achieve acceptable mesh quality.

\paragraph{Hybrid scene representation.}
To overcome the limitations of SDF methods in capturing fine structures, recent approaches integrate volumetric and SDF representations by dividing scenes into distinct regions and applying specialized reconstruction or sampling strategies. \citet{turki2024hybridnerf} and \citet{wang2023adaptive} segment scenes based on the scaling factor used to convert SDF into density, and employ distinct sampling techniques for volumetric and surface regions to enable real-time, high-quality rendering. StreetSurf~\cite{streetsurf} models different regions -- \textit{close, far, and sky} -- using different hierarchical space partitioning (e.g., 3D/4D hash grids, occupancy grids). This improves performance in urban scenes but imposes strong priors tied to vehicle ego poses, limiting generalizability. 
Meanwhile, ViiNeuS~\cite{scilla} initializes SDF sampling with volumetric density before gradually transitioning to a pure SDF representation, accelerating convergence and achieving state-of-the-art results on autonomous driving benchmarks. However, this approach risks introducing artifacts from the volumetric representation that can adversely affect the final SDF reconstruction.

Building on these insights, we propose a more adaptive divide-and-conquer technique that dynamically partitions the scene based on uncertainty across both representations. By applying tailored sampling and regularization strategies for different region, our method preserves fine details while effectively handling the large planar areas characteristic of autonomous driving environments.

\section{Method}
\begin{figure*}[ht]
    \vspace{-0.5 cm}
    \centering 
    \includegraphics[width=0.98\textwidth]{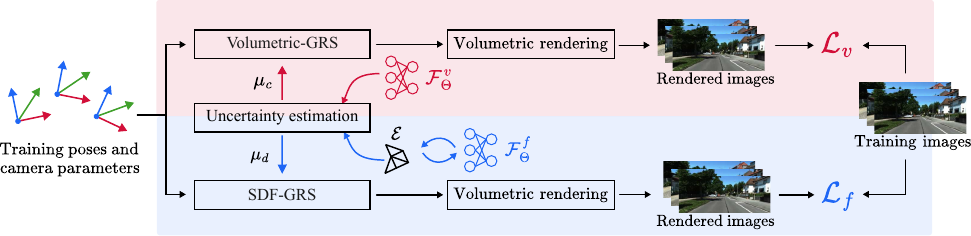} 
\caption{\textbf{Overview of \method}: we jointly train two implicit model: a volumetric representation $\mathcal{F}_\Theta^v$ and a SDF $\mathcal{F}_\Theta^f$ with a mutual guidance provided through our Guided Ray Sampling strategy. A colorized mesh of the scene $\mathcal{E}$ is periodically extracted from the SDF representation.}
    \vspace{-0.5 cm}
    \label{fig:method_overview}
\end{figure*}

Given a collection of RGB images captured from a moving vehicle in an urban area with limited overlap, our goal is to resolve perspective geometry ambiguities introduced by partial scene observations and to accurately reconstruct surfaces with precise structural details. To achieve this, we propose selectively fusing volumetric and SDF representations, with uncertainty guiding the division of labor between the two representations. Specifically, we use two implicit representations: a volumetric radiance field $\mathcal{F}_\Theta^v$ and an SDF field $\mathcal{F}_\Theta^f$
with trainable weights $\Theta_v$ and $\Theta_f$, respectively. We also extract at regular intervals during training a colorized mesh $\mathcal{E}$ from $\mathcal{F}_\Theta^f$ using the marching cubes algorithm. An overview of our method is presented in \cref{fig:method_overview}

To selectively fuse NeRF and SDF representations during training, we first introduce photometric and geometric uncertainty estimation across the two representations $\mathcal{F}_\Theta^v$ and $\mathcal{F}_\Theta^f$ (Section.~\ref{sec:uncertainty}). We then jointly optimize both representations with Guided Ray Sampling based on uncertainty estimation to leverage the strengths of both representations (Section.~\ref{sec:grs}). Additionally, to ensure the preservation of fine structures, we relax the Eikonal constraint and monocular supervision in uncertain regions to avoid over-regularization of the SDF field (Section.~\ref{sec:adaptive_relaxation}).

\subsection{Cross Representation Uncertainty Estimation}
\label{sec:uncertainty}
Due to the inherent partial observations in autonomous driving scenarios, both volumetric and SDF representations exhibit epistemic uncertainty in regions where visual cues are sparse. These regions are characterized by high variance in RGB and depth predictions, or deviations from the ground truth. To fully leverage the strengths of both representations, it is crucial to identify areas with high uncertainty and adaptively apply tailored sampling and regularization strategies. In the following sections, we first introduce the key concepts and notation related to our volumetric and SDF representations. We then describe how uncertainty is estimated to selectively fuse both representations effectively.

\paragraph{Background - Implicit volumetric representation.}
A volumetric radiance field is a continuous function $\mathcal{F}_\theta^v$ mapping a position and direction pair$(\mathbf{x},\mathbf{u)} \in \mathbb{R}^3 \times \mathbb{S}^2$ to a volume density $\sigma \in \mathbb{R}^+$ and a color $\mathbf{c} \in [0,1]^3$. \citet{2020nerf} model this function with a multi layer perceptron (MLP) whose weights $\theta$ are optimized to reconstruct a 3D scene given a set of posed images during training.

To render an image and depth, volume rendering is applied to alpha-composite the color for each ray, yielding the final pixel color $ \hat{C}_{v}(\mathbf{r}) \in \mathbb{R}^3 $ and the depth value $ \hat{D}_{v}(\mathbf{r}) \in \mathbb{R}^{+} $ to be:

\begin{equation}
    \hat{C}_{v}(\mathbf{r}) = \sum_{i=1}^{N} w_i c_i,  \ \  w_i = T_i \alpha_i, 
    \label{eq:volume-rendering}
\end{equation}

\begin{equation}
    \hat{D}_{v}(\mathbf{r}) = \sum_{i=1}^{N} w_i z_i,  \ \  w_i = T_i \alpha_i, 
    \label{eq:depth-volume-rendering}
\end{equation}
where $T_i = \prod_{j=1}^{i-1} (1- \alpha_j)$ is the accumulated transmittance, $\alpha_i \in \mathbb{R}$ the blending factor and $z_i \in \mathbb{R}^{+}$ is the distance of the sample to the camera center. Here, $\alpha_i$ is computed from the predicted density: 
$\alpha_i = 1 - \exp (-\sigma_i \delta_i )$,
with $\delta_i \in \mathbb{R}^{+}$ being the distance between samples along the ray.

\paragraph{Background - Implicit surface representation.}
While the volumetric field predicts a pointwise 3D density $\sigma(x)$ at each location $x$, the SDF instead predicts a signed distance function $f(x)$ and converts it to density using a logistic function $\phi_s(f)$ with a global scale parameter $s$, effectively confining the density to a narrow band of width $O(1/s)$ around the surface~\cite{wang2023adaptive}. To enable volume rendering, the signed distance function $f$ is then transformed into the density $\sigma$ using sigmoid-shape mapping for alpha compositing. NeuS~\cite{NeuS} adopted a new formulation of the blending factor:
\begin{equation}
    \alpha_i = \text{max} \left(\frac{ \Phi_s(f(p_i)) - \  \Phi_s(f(p_{i+1}))}{ \Phi_s(f(p_i))}, 0\right),
    \label{eq:neus-alpha}
\end{equation}
\noindent where $f(p_i)$ and $f(p_{i+1})$ are signed distance values at section points centered on $x_i$, $\Phi_s(x)$ the sigmoid function.

\paragraph{Uncertainty estimation.}

When reconstructing urban scenes, the direct 3D point-based density prediction of the volumetric field enables rapid fitting of high-frequency geometry (e.g., poles), but may also introduce spurious density (floaters) on large planar surfaces~\cite{NeuS}. In contrast, the more constrained density prediction of the SDF promotes surface continuity, but may over-smooth fine details when using a uniform or overly small scale parameter $s$. To harness the complementary strengths of both representations, it is essential to identify and suppress their unreliable predictions. To this end, we introduce two complementary uncertainty estimates: \textbf{geometric uncertainty} ($\mu_d$) and \textbf{photometric uncertainty} ($\mu_c$).

The first type of uncertainty focuses on identifying regions with ambiguous geometric information: areas where visual cues are insufficient for geometry reconstruction (\textit{e.g.}, textureless surfaces or partially observed complex structures). Ideally, one would measure epistemic uncertainty by comparing predictions with dense 3D ground truth. However, in autonomous driving scenarios, such ground truth is often unavailable. Therefore, we propose to measure uncertainty by evaluating the consistency between the two representations: Given a ray $r$, we introduce the geometry uncertainty based on the rendered depth from $\mathcal{F}_\Theta^v$, and the distance to the first intersection with the mesh $\mathcal{E}$, $\hat{D}_{\mathcal{E}}(\mathbf{r})$:

\begin{equation}
    \mathcal{\mu}_{d}(\mathbf{r})= |1 - \frac{\hat{D}_{\mathcal{E}}(\mathbf{r})}{\hat{D}_{v}(\mathbf{r})}|.
\end{equation}
High values of $\mu_{d}$ indicate inconsistency between the volumetric and SDF models, suggesting that at least one representation is uncertain. In such cases, we preferentially rely on the volumetric field for two key reasons. First, during initialization, it captures fine structures more completely and achieves faster convergence~\cite{scilla}. Second, in later stages, probability sampling~\ref{sec:proba_sampling} of $\mathcal{F}_\Theta^f$ can effectively filter out noise introduced by the volumetric field $\mathcal{F}_\Theta^v$ while preserving the fine details that $\mathcal{F}_\Theta^v$ initially localized.

\begin{figure}[!b] 
\small
\centering

\def\fgsize{0.45}
\def\rowspacing{0.05cm}

\scriptsize
\setlength{\tabcolsep}{0.0035\linewidth}
\renewcommand{\arraystretch}{1.0}
\vspace{-0.8cm}
\begin{tabular}{ccc}

& $\mu_d$ & $\mu_c$\\ 

 \vspace{\rowspacing}
    \multirow{1}{*}[8 mm]{\rotatebox[origin=c]{90}{Epoch~01}}  &
    \includegraphics[clip=false, trim={0 0 0 0},width= \fgsize\columnwidth]{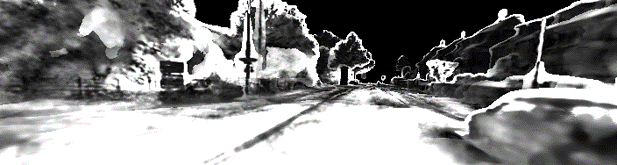} & 
    \includegraphics[clip=false, trim={0 0 0 0},width= \fgsize\columnwidth]{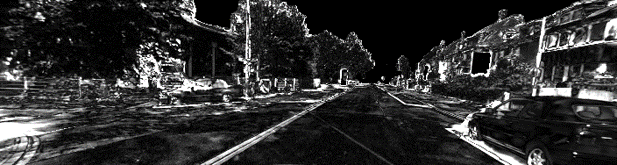}  \\

 \vspace{\rowspacing}
    \multirow{1}{*}[8 mm]{\rotatebox[origin=c]{90}{Epoch~11}}  &
    \includegraphics[clip=false, trim={0 0 0 0},width= \fgsize\columnwidth]{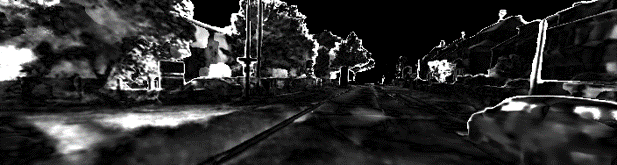} & 
    \includegraphics[clip=false, trim={0 0 0 0},width= \fgsize\columnwidth]{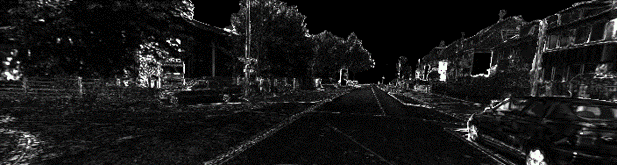}  \\

    \multirow{1}{*}[8 mm]{\rotatebox[origin=c]{90}{GT. RGB}}  & 
      \multicolumn{2}{c}{%
        \includegraphics[width=0.4\columnwidth]{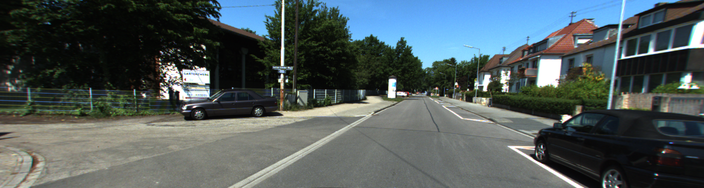}
      } 
      
\end{tabular}

\vspace{-0.4cm}
  \caption{\footnotesize Visualization of $\mu_d$ and $\mu_c$, maximum clamped at 0.3.}
  \vspace{-0.8cm}
\label{fig:vis_mu_dc}

\end{figure}

Secondly, we introduce a photometric-based uncertainty to evaluate the predictions of the SDF field. We hypothesize that if the learned SDF field is accurate, \textbf{a single-sample rendering} at the mesh depth should yield a precise photometric rendering. 
Casting a ray $\mathbf{r} = \mathbf{o} + t\mathbf{u}$ from the camera center $\mathbf{o}$ through the pixel along direction $\mathbf{u}$, we define the photometric uncertainty indicator as:
\begin{equation}
    \mathcal{\mu}_{c}(\mathbf{r})= |\hat{C}_{\mathcal{E}}(\mathbf{r}) - \bar{C}(\mathbf{r}) |,
    \label{equ: photometric}
\end{equation}
where $\bar{C}$ indicates rgb ground truth, and $\hat{C}_{\mathcal{E}}(\mathbf{r})$ represents the rgb value of $\mathcal{F}_\Theta^f$ at the point $\mathbf{o} + \hat{D}_{\mathcal{E}}(\mathbf{r})\mathbf{u} $, i.e. the point on the Mesh $\mathcal{E}$ of the given ray. 
$\mathcal{\mu}_{c}(\mathbf{r})$ is used to divide certain and uncertain regions in the SDF representation (see Section.~\ref{sec:adaptive_relaxation}). We use it to guide the sampling strategy of the $\mathcal{F}_{\theta}^{v}$ for certain regions, and adaptively relax geometric regularization in uncertain regions to prevent over-smoothing. While $\mu_c$ does not guarantee accurate depth alignment, it is sufficient to quickly populate large planar regions without requiring complete rendering of $\mathcal{F}_{\theta}^{v}$. We provide a visualization of $\mu_c$ and $\mu_d$ in Figure~\ref{fig:vis_mu_dc}. 

\subsection{Guided Ray Sampling (GRS)}
\label{sec:grs}

To selectively fuse the volumetric and SDF representation during training, we propose to jointly train $\mathcal{F}_\Theta^v$ and $\mathcal{F}_\Theta^f$ via Guided Ray Sampling (GRS) based on previously introduced uncertainty $\mathcal{\mu}_{c}$ and $\mathcal{\mu}_{d}$. Specifically, for the volumetric representation, we use $\mathcal{\mu}_{c}$ to guide the sampling toward the surface of the mesh $\mathcal{E}$ where SDF field $\mathcal{F}_\Theta^f$ is confident. For the SDF representation, 
we focus the sampling to the estimated surface of the best-suited representation, using geometric uncertainty $\mathcal{\mu}_d$.

\paragraph{Volumetric-GRS.}
We consider a batch of rays $r \in \mathcal{R}_v$ with for each ray $r$ the corresponding original sampling interval $\left[t_{n}, t_{f}\right]$. To improve volumetric reconstruction on planar surfaces, we adjust the sampling bounds to focus more on the estimated depth of the Mesh $D_\mathcal{E}(\mathbf{r})$ if the $\mathcal{\mu}_{c}(\mathbf{r})$ is below a certain threshold $\tau_c$:

\begin{equation}
   [t_{n}, t_{f}](\mathbf{r}) = 
\begin{cases}
    \left[0, D_\mathcal{E}(\mathbf{r}) + \delta\right] & \text{if } \mathcal{\mu}_{c} < \tau_c, \\
    \left[0, \infty \right[          & \text{otherwise, }
\end{cases}
\end{equation}
where $\delta$ is a hyperparameter (analogous to the shell factor in~\cite{wang2023adaptive}) periodically updated. This helps avoid noisy planar predictions that may arise from the under-constrained optimization of the volumetric representation, ensuring more accurate surface reconstruction. In practice, $\tau_c$ is set as a constant for each dataset (see Section. 2 in supplementary material).

\paragraph{SDF-GRS.}

After rendering $\mathcal{F}_\Theta^v$ with guided sampling, the predicted depth $\hat{D}_{v}(\mathbf{r})$ becomes available, allowing us to estimate the geometric uncertainty $\mathcal{\mu}_{d}$ to direct the sampling of $\mathcal{F}_\Theta^f$. Since high $\mathcal{\mu}_{d}$ values indicates regions where the predictions of the two representations are inconsistent, it is crucial to decide which representation to trust. Here, we propose trusting the volumetric representation, as it typically learns faster and captures complex structures more effectively and expect that noisy predictions from $\mathcal{F}_\Theta^v$  will be handled by the Volumetric-GRS.

Similar to Volumetric-GRS, we adjust the sampling bounds based on the geometric uncertainty $\mathcal{\mu}_d(\mathbf{r})$. If $\mathcal{\mu}_d(\mathbf{r})$ exceeds a threshold $\tau_d$, we focus the sampling on the estimated depth from the \textbf{volumetric field} $\hat{D}_{v}(\mathbf{r})$. Otherwise, sampling is concentrated around the depth of the mesh $D_\mathcal{E}(\mathbf{r})$, ensuring accurate surface reconstruction in both cases:

\begin{equation}
   [t_n, t_f](\mathbf{r}) = 
\begin{cases}
    [D_\mathcal{E}(\mathbf{r}) - \delta, D_\mathcal{E}(\mathbf{r}) + \delta] & \text{if }\mathcal{\mu}_{d} < \tau_d,\\
    [D_v(\mathbf{r}) - \delta, D_v(\mathbf{r}) + \delta]& \text{otherwise,}
\end{cases}
\end{equation}
with $\delta$ being the same shell factor hyperparameter described in the previous paragraph.

\paragraph{Adaptive thresholding.}
\label{sec:adaptive-unceratinty}
The SDF-GRS strategy employs a threshold $\tau_{d}$ to classify rays as “certain” or “uncertain,” but selecting an appropriate value a priori can be difficult across varying scenes and reconstruction tasks. To overcome this, we propose a lightweight, data‐driven algorithm that adaptively adjusts $\tau_{d}$ based on the observed ratio of certain to uncertain rays (see Algorithm~\ref{alg:adaptive-tau}). Empirically, this adaptive scheme generalizes robustly across diverse urban environments (see supplementary material for details).  

\begin{algorithm}[htbp]
\footnotesize
\caption{\footnotesize Adaptive Uncertainty Threshold $\tau$}
\label{alg:adaptive-tau}
\KwIn{$\tau$, $\{\mu_d(r_i)\}_{i=1}^N$: depth uncertainties of $N$ rays}\
\KwData{$\gamma_{\uparrow},\gamma_{\downarrow}$: growth/decay factors; 
$\rho_{\text{high}},\rho_{\text{low}}$: ratio thresholds}\
\KwOut{Updated $\tau$}
$u \leftarrow \sum_i[\mu_d(r_i) > \tau]$;
$c \leftarrow (N - u)$;
$\rho \leftarrow (u / c)$\;
\If{$\rho > \rho_{\text{high}}$}{
    $\tau \leftarrow \tau \times \gamma_{\uparrow}$\;
}
\ElseIf{$\rho < \rho_{\text{low}}$}{
    $\tau \leftarrow \tau \times \gamma_{\downarrow}$\;
}
\Return $\tau$
\end{algorithm}

\subsection{Adaptive Relaxation on Geometric Constrain}
\label{sec:adaptive_relaxation}
Another critical factor in capturing fine structures within the SDF field is avoiding over-regularization while the geometry remains under-optimized~\cite{wang2024raneus, wang2024neurodin}. Whereas previous urban reconstruction methods enforce the Eikonal constraint across the entire scene~\cite{scilla, streetsurf}, we adaptively relax it in regions where the SDF prediction is uncertain.
Additionally, we relax normal supervision in the same way based on pseudo ground truth normals $\bar{N}(r)$ predicted by an off-the-shelf network~\cite{eftekhar2021omnidata}, circumventing unreliable supervision. Below, we define our two uncertainty-aware geometric regularization term:
\begin{multline}
\label{equ:normal}
\mathcal{L}_{N}^{u} = \mathbb{I}_{\mu_c} \cdot \Biggl(
    \Bigl\|\frac{\nabla f(x_N)}{\|\nabla f(x_N)\|_2} - \bar{N}(r)\Bigr\| \;+\; \\
    \Bigl\|1 - \left(\frac{\nabla f(x_N)}{\|\nabla f(x_N)\|_2}\right)^{\top} \bar{N}(r)\Bigr\|
\Biggr),
\end{multline}

\begin{equation}
    \mathcal{L}_{\text{eik}}^{u} = 
    \mathbb{I}_{\mu_c} \cdot \left( \left\| \nabla f(x_{(i,j)}) \right\|_2 - 1 \right)^2,
\end{equation}
\noindent where \(x_N\) denotes the closest sample to the surface, as described in \cite{scilla}.
We define the indicator function \(\mathbb{I}_{\mu_c}\) as follows, which can be regarded as a trimmed robust kernel that down-weights uncertain regions during supervised loss computation, in line with RobustNeRF~\cite{sabour2023robustnerf}:

\[
\mathbb{I}_{\mu_c} \;=\;
\begin{cases}
1, & \text{if } \mu_c(\mathbf{r}) > \tau_c,\\[6pt]
0, & \text{otherwise}.
\end{cases}
\]

\subsection{Optimization details}
\paragraph{Probability sampling.}
\label{sec:proba_sampling}
While the GRS mechanism adjusts the sampling boundaries of each ray based on uncertainty, we further refine the sampling process using a density estimator. Specifically, we employ a proposal network -- \textit{similar to ViiNeuS~\cite{scilla}} -- that is self-supervised by the volumetric field $\mathcal{F}_\Theta^v$ using the proposal loss $\mathcal{L}_p$ introduced in MipNeRF360~\cite{mip-nerf-360}. At the end of the training, we switch to computing the PDF weights directly from $\mathcal{F}_\Theta^f$ to allow the SDF representation to be freely refined.

\paragraph{Losses.}
For both representations, \( \mathcal{F}_\Theta^v \) and \( \mathcal{F}_\Theta^f \), we employ a standard \( L_1 \) loss to minimize the pixel-wise color difference between the rendered image \( \hat{C} \) and the ground truth image \( \bar{C} \), along with a DSSIM~\cite{wang2023planerf} loss on color patches. These losses are jointly denoted as \( \mathcal{L}_{\text{rgb}} \). Similar to StreetSurf~\cite{streetsurf}, we model the sky color using an auxiliary MLP conditioned on the ray direction and introduce a sky loss \( \mathcal{L}_{\text{sky}} \) to enforce zero opacity for sky pixels. The segmentation mask is generated using an off-the-shelf semantic segmentation network~\cite{cheng2021mask2former}. In addition to the geometric regularization described in Section.~\ref{sec:adaptive_relaxation}, we apply distortion regularization \( \mathcal{L}_{\text{d}} \) from MipNeRF~\cite{mip-nerf-360} to mitigate floaters. Finally, to enhance fine structure learning, we incorporate an additional semantic head into \( \mathcal{F}_\Theta^v \) and impose a cross-entropy loss for semantic supervision, denoted as \( \mathcal{L}_{\text{sem}} \).

The total losses for the volumetric field ($\mathcal{L}_{\text{v}}$) and the SDF field ($\mathcal{L}_{\text{f}}$) are:
{\small
\begin{align}
    \mathcal{L}_{\text{v}} = 
    \mathcal{L}_{\text{rgb}} 
    + \lambda_d \mathcal{L}_d 
    + \lambda_{\text{sky}} \mathcal{L}_{\text{sky}} 
    + \lambda_{\hat{N}} \mathcal{L}_{\hat{N}}
    + \lambda_{\text{sem}} \mathcal{L}_{\text{sem}}, \\
    \mathcal{L}_{\text{f}} = 
    \mathcal{L}_{\text{rgb}} 
    + \lambda_d \mathcal{L}_d 
    + \lambda_{\text{sky}} \mathcal{L}_{\text{sky}} 
    + \lambda_{\hat{N}} \mathcal{L}_{N}^{u}
    + \lambda_{\text{eik}} \mathcal{L}_{\text{eik}}^{u},
\end{align}
}where $\lambda_d$, $\lambda_{\text{sky}}$, $\lambda_{\text{normal}}$, $\lambda_d$, $\lambda_{\text{normal}}$, $\lambda_{\text{eik}}$ are scaling factors. During the early epochs, we reduce the coefficients for both normal supervision and distortion regularization to facilitate a stable initialization.

\section{Experiments}
\paragraph{Implementation details}
We use hash encoding to encode the positions~\cite{mueller2022instant}, and spherical harmonics to encode the viewing directions. We use 2 layers with 64 hidden units for the MLPs $\mathcal{F}_\Theta^h$ and $\mathcal{F}_\Theta^c$. We trained our model on a single consumer GPU with 24GB of VRAM, using Adam optimizer with a cosine learning rate decay from $10^{-2}$ to $10^{-4}$.
We use Marching Cubes~\cite{lorensen1998marching} to generate the final mesh that represents the scene. Further implementation details can be found in the supplementary materials.

\begin{table*}[tb]
    \centering
    \resizebox{1.7\columnwidth}{!}{%
    \begin{tabular}{lcccccccccccccccccc}
    
    & \multicolumn{8}{c}{KITTI-360~\cite{Kitti}}  
    & & \multicolumn{8}{c}{Pandaset~\cite{pandaset}}\\
    \cmidrule[\lightrulewidth]{2-9}
    \cmidrule[\lightrulewidth]{11-18}
    & \multicolumn{2}{c}{Seq. 30} & \multicolumn{2}{c}{Seq. 31} & \multicolumn{2}{c}{Seq. 35} & \multicolumn{2}{c}{Seq. 36} & & 
    \multicolumn{2}{c}{Seq. 23} & \multicolumn{2}{c}{Seq. 37} & \multicolumn{2}{c}{Seq. 42} & \multicolumn{2}{c}{Seq. 43}\\
    \cmidrule(lr){2-3} \cmidrule(lr){4-5} \cmidrule(lr){6-7}\cmidrule(lr){8-9}
    \cmidrule(lr){11-12} \cmidrule(lr){13-14} \cmidrule(lr){15-16}\cmidrule(lr){17-18}
    & P$\rightarrow$M & Prec.  & P$\rightarrow$M & Prec. & P$\rightarrow$M & Prec. & P$\rightarrow$M~ & Prec. &
    & P$\rightarrow$M & Prec. & P$\rightarrow$M & Prec. & P$\rightarrow$M & Prec. & P$\rightarrow$M & Prec.  \\
    \cmidrule[\lightrulewidth]{1-9}
    \cmidrule[\lightrulewidth]{11-18}

    StreetSurf~\cite{streetsurf} & 0.14 & 0.50 & \cellcolor{sdf} \textbf{0.09} & 0.71 &  \underline{0.10} &   0.67 &  \underline{0.11}  & 0.66 & & 2.52 & 0.17 & 0.25 & \cellcolor{sdf} \textbf{0.66} & 0.36 &   0.29& \underline{0.19} & 0.29 \\
    ViiNeuS~\cite{scilla} & \underline{0.13} &  0.56 &  \underline{0.11} &  0.71 & 0.11 & 0.66 & 0.13 & 0.72 & & \cellcolor{sdf} \textbf{0.17}&  \underline{0.35} & \cellcolor{sdf} \textbf{0.22} &   0.44 & \cellcolor{sdf} \textbf{0.15} &  0.59 & \cellcolor{sdf} \textbf{0.17} &  0.45\\

    GOF~\cite{Yu2024GOF} - sparse &  -- & -- &  0.29 & 0.53&    0.23 &  0.63&  -- &  -- & &  0.45 & 0.32 & 0.28 &  0.60 &  0.28&   0.46&   0.50 &  0.35 \\

    GOF~\cite{Yu2024GOF} - dense &  0.17 & \underline{0.71} &  0.16 &  \underline{0.72} &    0.20 &   \underline{0.74} &   \underline{0.11} & \underline{0.80} & &  0.37 &  \underline{0.35} & 0.27 & \underline{0.62} &  0.48 & 0.33&   0.31 & 0.44 \\
    
    \cmidrule[\lightrulewidth]{1-9}
    \cmidrule[\lightrulewidth]{11-18}

    \method~(ours) & \cellcolor{sdf}  \textbf{0.10} &    \cellcolor{sdf}  \textbf{0.78} & \cellcolor{sdf} \textbf{ 0.09} & \cellcolor{sdf}  \textbf{ 0.85} & \cellcolor{sdf} \textbf{0.09} & \cellcolor{sdf} \textbf{ 0.84} & \cellcolor{sdf} \textbf{0.09} & \cellcolor{sdf} \textbf{0.85} & &\underline{0.21} & \cellcolor{sdf} \textbf{0.38} &\underline{0.23} &  0.53 & \underline{0.20} & \cellcolor{sdf} \textbf{0.62} & 0.20 &  \cellcolor{sdf} \textbf{0.52}\\
    
    \cmidrule[\lightrulewidth]{1-9}
    \cmidrule[\lightrulewidth]{11-18}

    \\
    & \multicolumn{8}{c}{Waymo~\cite{Waymo}}  
    & & \multicolumn{8}{c}{nuScenes~\cite{nuscenes}}\\
    \cmidrule[\heavyrulewidth]{2-9}
    \cmidrule[\heavyrulewidth]{11-18}
    & \multicolumn{2}{c}{Seq. 10061} & \multicolumn{2}{c}{Seq. 13196} & \multicolumn{2}{c}{Seq. 14869} & \multicolumn{2}{c}{Seq. 102751} & &    
    \multicolumn{2}{c}{Seq. 0034} & \multicolumn{2}{c}{Seq. 0071} & \multicolumn{2}{c}{Seq. 0664} & \multicolumn{2}{c}{Seq. 0916}\\
    \cmidrule(lr){2-3} \cmidrule(lr){4-5} \cmidrule(lr){6-7}\cmidrule(lr){8-9}
    \cmidrule(lr){11-12} \cmidrule(lr){13-14} \cmidrule(lr){15-16}\cmidrule(lr){17-18}
    & P$\rightarrow$M & Prec.  & P$\rightarrow$M & Prec. & P$\rightarrow$M & Prec. & P$\rightarrow$M~ & Prec. &
    & P$\rightarrow$M & Prec. & P$\rightarrow$M & Prec. & P$\rightarrow$M & Prec. & P$\rightarrow$M & Prec.  \\
    \cmidrule[\lightrulewidth]{1-9}
    \cmidrule[\lightrulewidth]{11-18}

    StreetSurf~\cite{streetsurf} &  \underline{0.22} & 0.43 & 0.35 & \cellcolor{sdf}\textbf{0.53} & 0.23 & 0.35 & 0.25 & 0.24 & & 0.57 & \underline{0.29} & 0.78 & 0.47 & 0.67 & \underline{0.50} & 0.65 & 0.28 \\
    ViiNeuS~\cite{scilla}& \cellcolor{sdf} \textbf{0.19}& \underline{0.44} &\cellcolor{sdf} \textbf{0.22} & \underline{0.48} &\cellcolor{sdf} \textbf{0.14}& \underline{0.47} & \cellcolor{sdf}\textbf{0.19}&  0.30 & & \underline{0.40} & 0.20 &\cellcolor{sdf}\textbf{0.22} &\cellcolor{sdf}\textbf{0.59} & \underline{0.40} & 0.40 & \underline{0.22} & \underline{0.54}\\

    GOF~\cite{Yu2024GOF} - sparse & 1.87 & 0.32 & 2.32 & 0.20 & 1.63 & 0.36 & 1.54 & 0.29 & & 1.55 & 0.07 & 1.72 & 0.16 &1.49 & 0.12 & 1.41 & 0.18 \\

    GOF~\cite{Yu2024GOF} - dense & 1.20 & 0.38 & 1.17 & 0.39 & 1.55 & 0.41 & 2.11 & \underline{0.34} & & 1.02 & 0.12 & 1.55 & 0.23 &1.44 & 0.12 & 1.06 & 0.29 \\
    \cmidrule[\lightrulewidth]{1-9}
    \cmidrule[\lightrulewidth]{11-18}

    \method~(ours) & 0.23 & \cellcolor{sdf}\textbf{0.47} & \underline{0.27} & \underline{0.48} & \underline{0.19} & \cellcolor{sdf} \textbf{0.60} & \underline{0.22}&   \cellcolor{sdf} \textbf{0.46} & &    \cellcolor{sdf} \textbf{0.30} &  \cellcolor{sdf} \textbf{0.43} & \underline{0.35} & \underline{0.56} &  \cellcolor{sdf} \textbf{0.26} &   \cellcolor{sdf} \textbf{0.56} &  \cellcolor{sdf} \textbf{0.21}&    \cellcolor{sdf} \textbf{0.64}\\
    
    \cmidrule[\lightrulewidth]{1-9}
    \cmidrule[\lightrulewidth]{11-18}
        
  \end{tabular}
    }
    \caption{Quantitative results on KITTI-360 \cite{Kitti}, Pandaset \cite{pandaset}, Waymo Open Dataset \cite{Waymo} and nuScenes \cite{nuscenes}. We report the mean Point to Mesh (P$\rightarrow$M) distance in meters $m$, and the percentage of points with a distance to mesh below 0.15$m$ (Prec.). We highlight best performing methods in \colorbox{sdf}{\textbf{green}} and second one \underline{underlined}. Missing entry (--) designate failure case.
    }
    \vspace{-0.3cm}
    \label{tab:quantitative_results}
\end{table*}

\begin{figure}[t]
    \begin{minipage}{0.98\linewidth}
        \centering
        \resizebox{\linewidth}{!}{%
            \begin{tabular}{lcc cccccc}
                \cmidrule[\lightrulewidth]{2-9}
                & \multicolumn{2}{c}{KITTI} & \multicolumn{2}{c}{Pandaset} & \multicolumn{2}{c}{Waymo} & \multicolumn{2}{c}{nuScenes} \\ 
                \cmidrule(lr){2-3} \cmidrule(lr){4-5} \cmidrule(lr){6-7} \cmidrule(lr){8-9}
                & PSNR & SSIM & PSNR & SSIM & PSNR & SSIM & PSNR & SSIM \\ 
                \cmidrule[\lightrulewidth]{1-9} 
                StreetSurf       & 24.04            & 0.83            & 22.24           & 0.66         & 23.42 & 0.77  & \underline{22.28}& 0.76               \\ 
                ViiNeuS          & \cellcolor{sdf} \textbf{24.83}   & \cellcolor{sdf} \textbf{0.89}         & 22.95           & 0.80      & \underline{23.74} & \underline{0.87}      & 21.96            & \underline{0.83}  \\ 
                GOF              & 23.34            & 0.86            & \cellcolor{sdf} \textbf{25.54}  & \cellcolor{sdf} \textbf{0.87}  & 20.92            & 0.81              & 19.13 & 0.76 \\
                \cmidrule[\lightrulewidth]{1-9}
                \method~(ours)& \underline{24.11}& \underline{0.88}& \underline{23.1}& \underline{0.80}& \cellcolor{sdf} \textbf{24.90} & \cellcolor{sdf}  \textbf{0.88} & \cellcolor{sdf} \textbf{24.10} & \cellcolor{sdf}  \textbf{0.86}    \\
                \cmidrule[\lightrulewidth]{1-9}
            \end{tabular}
        }
        \vspace{-0.4cm}
        \captionsetup{type=table}
        \caption{\footnotesize Mean photometric results for each dataset. We highlight best performing methods in \colorbox{sdf}{\textbf{green}} and second one \underline{underlined}.}
        \label{tab:photometric_res}
    \end{minipage}
    \vspace{-5mm}
\end{figure}

\paragraph{Datasets}
We evaluate our method on four public driving datasets: KITTI-360~\cite{Kitti}, nuScenes~\cite{nuscenes}, Waymo Open Dataset~\cite{Waymo}, and Pandaset~\cite{pandaset}. For each dataset, we select four diverse scenes to capture a wide range of urban settings. In the case of Pandaset, we focus on static sequences and mask dynamic vehicles to ensure consistent evaluation.
\paragraph{Baselines}
We compare our proposed solution to current state-of-the-art (SoTA) surface reconstruction methods, including the SDF-based approaches StreetSurf \cite{streetsurf} and ViiNeuS \cite{scilla}, as well as the Gaussian splatting-based method GoF \cite{Yu2024GOF}.
StreetSurf~\cite{streetsurf} models close, far, and sky regions using hierarchical space partitioning, with 3D and 4D hash grids and occupancy grids for efficient ray sampling. ViiNeuS~\cite{scilla} initializes the SDF field with volumetric density predictions, achieving SoTA results in autonomous driving scenario. GoF~\cite{Yu2024GOF} is a Gaussian splatting method that achieves SoTA performance in object-centric scenes with high image overlap. It leverages 2D Gaussian splatting regularization losses~\cite{huang20242d} and introduces an enhanced mesh extraction solution tailored for 3D Gaussian splatting (3DGS). In our experiments, we initialize GoF using COLMAP-derived sparse and dense point clouds, denoted as GoF-sparse and GoF-dense, respectively.

\paragraph{Evaluation metrics}
In addition to conventional photometric metrics (see Table~\ref{tab:photometric_res}), we assess the quality of the reconstructed meshes using two metrics similar to those in ViiNeuS~\cite{scilla}. 
\begin{itemize}
    \item \noindent \textbf{Point to Mesh (P$\rightarrow$M):} the mean distances from the ground truth LiDAR points to the predicted SDF-generated mesh.
    \item \noindent \textbf{Precision (Prec.):} the percentage of LiDAR points with a distance to the mesh below $0.15$m.
\end{itemize}

\subsection{Results}
\paragraph{Quantitative analysis.}
\label{sec:quantitative}
We report quantitative results across four datasets in Tab.~\ref{tab:quantitative_results}. Our method consistently outperforms or matches SoTA approaches, achieving the top metrics in most scenes on KITTI-360 and nuScenes and delivering competitive results on Pandaset and Waymo.

We observe higher P$\rightarrow$M errors on Waymo and PandaSet due to densely occluded vegetation present in the LiDAR ground truth, not completely visible in the training images (see Figure.~2 in the supplementary). Our GRS strategy focuses sampling around the “visible” surface, forming a thin shell around trees and shrubs, which inflates the P$\rightarrow$M error. Despite this, Our method obtains the best photometric score on Waymo and the highest average precision on each individual dataset. Averaged over the four datasets, our method achieves a precision score of 0.60, outperforming ViiNeuS (0.49) and other methods. Additionally, we attain the best mean P$\rightarrow$M error of 0.20 m, equivalent to ViiNeuS, with StreetSurf at 0.47 m and GoF-dense at 0.82 m. These results highlight our strength in capturing precise geometry, underscoring the effectiveness of our joint optimization design.

\begin{figure}
\centering
\begin{subfigure}{0.3\linewidth}
    \centering
    \includegraphics[width=1.0\textwidth]{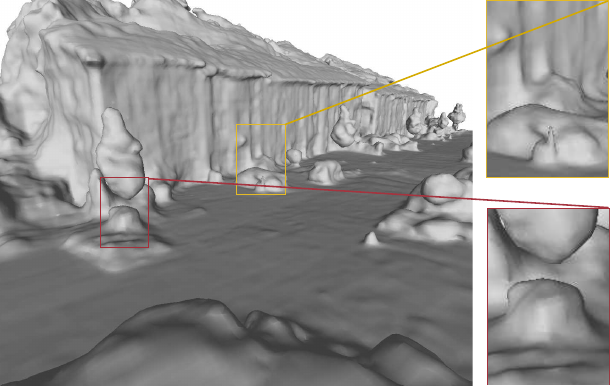} 
    \caption{\footnotesize w/o GRS}
\label{fig:abl_no_guided_sampling}
\end{subfigure}
\begin{subfigure}{0.3\linewidth}
    \centering
    \includegraphics[width=1.0\textwidth]{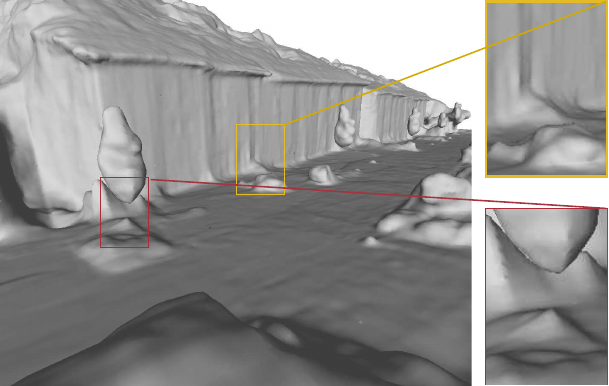} 
    \caption{\footnotesize w/o Ada. Relax.} 
    \label{fig:abl_no_masked_reg}
\end{subfigure}
\begin{subfigure}{0.3\linewidth}
    \centering
    \includegraphics[width=1.0\textwidth]{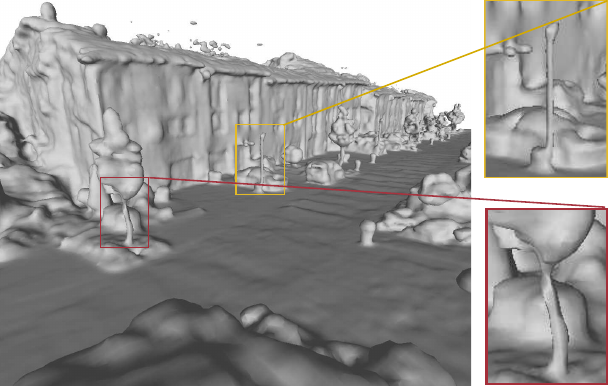} 
    \caption{\footnotesize \method~(ours)} 
    \label{fig:abl_ours}
\end{subfigure}\\
\vspace{-2mm}
\caption{Ablations study: we deactivate some \method~'s key components. (a) without GRS directed by $\mathcal{\mu}_{d,c}$  (b) without adaptive relaxation on Eikonal constrain and normal supervision \label{fig:ablation}}
\vspace{-5mm}
\end{figure}

\begin{figure*}[tb] 

\centering

\def\fgsize{0.44}
\def\rowspacing{0.1cm}

\scriptsize
\setlength{\tabcolsep}{0.0035\linewidth}
\renewcommand{\arraystretch}{1.0}
\begin{tabular}{ccccc}

      & StreetSurf~\cite{streetsurf} & ViiNeuS~\cite{scilla} & GoF~\cite{Yu2024GOF}-dense & \method~(ours)\\ 
    
 \vspace{\rowspacing}
    \multirow{1}{*}[31mm]{\rotatebox[origin=c]{90}{KITTI-360~\cite{Kitti}~\textendash~Seq.~30}}  &
    \includegraphics[clip=false, trim={0 0 0 0},width= \fgsize\columnwidth]{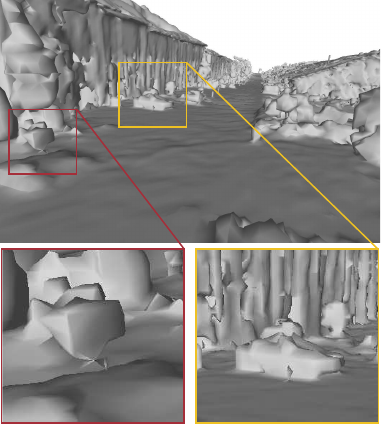} & 
    \includegraphics[clip=false, trim={0 0 0 0},width= \fgsize\columnwidth]{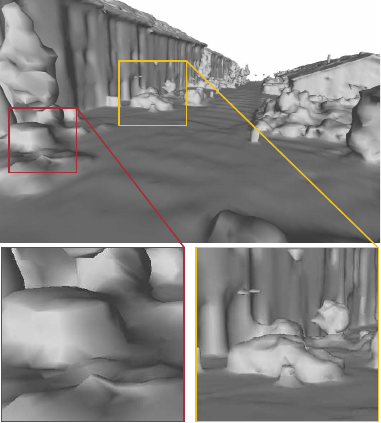} & 
    \includegraphics[clip=false, trim={0 0 0 0},width= \fgsize\columnwidth]{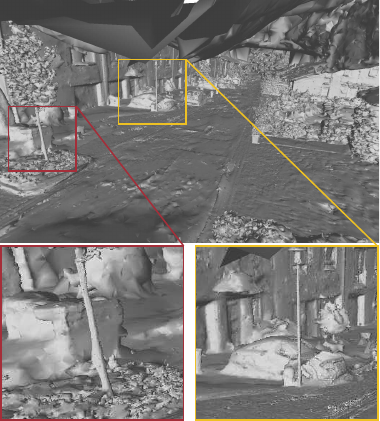} & 
    \includegraphics[clip=false, trim={0 0 0 0},width= \fgsize\columnwidth]{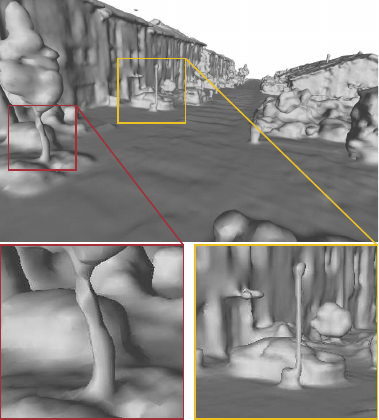} \\

    \vspace{\rowspacing}

    \multirow{1}{*}[33mm]{\rotatebox[origin=c]{90}{Pandaset~\cite{pandaset}~\textendash~Seq.~037}}  &
    \includegraphics[clip=false, trim={0 0 0 0},width= \fgsize\columnwidth]{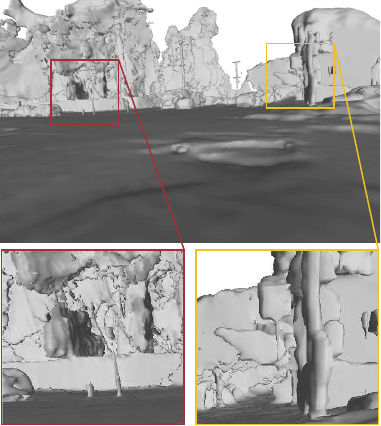} & 
    \includegraphics[clip=false, trim={0 0 0 0},width= \fgsize\columnwidth]{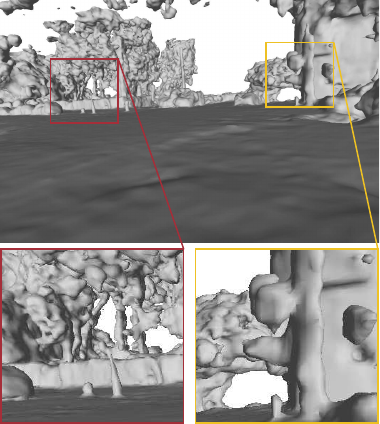} & 
    \includegraphics[clip=false, trim={0 0 0 0},width= \fgsize\columnwidth]{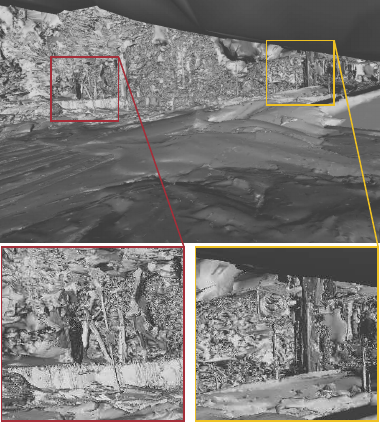} & 
    \includegraphics[clip=false, trim={0 0 0 0},width= \fgsize\columnwidth]{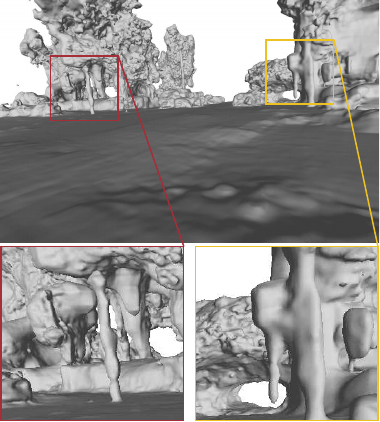} \\

 \vspace{\rowspacing}

    \multirow{1}{*}[34mm]{\rotatebox[origin=c]{90}{Waymo~\cite{Waymo}~\textendash~Seq.~14689}}  &
    \includegraphics[clip=false, trim={0 0 0 0},width= \fgsize\columnwidth]{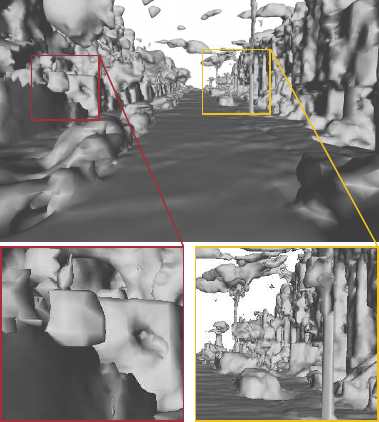} & 
    \includegraphics[clip=false, trim={0 0 0 0},width= \fgsize\columnwidth]{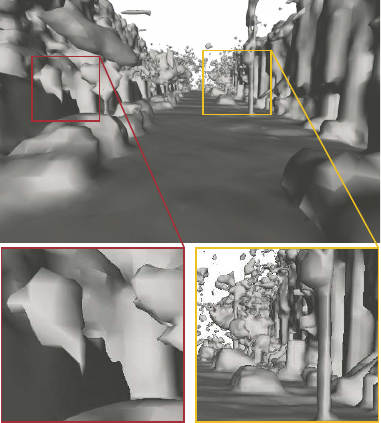} & 
    \includegraphics[clip=false, trim={0 0 0 0},width= \fgsize\columnwidth]{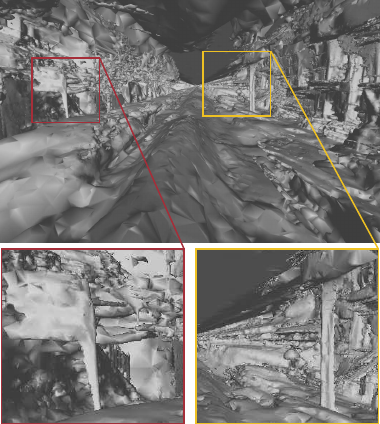} & 
    \includegraphics[clip=false, trim={0 0 0 0},width= \fgsize\columnwidth]{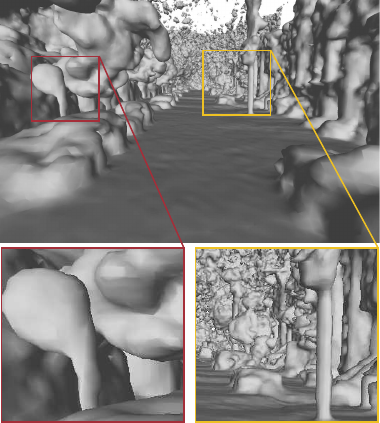} \\

     \vspace{\rowspacing}

    \multirow{1}{*}[33mm]{\rotatebox[origin=c]{90}{nuScenes~\cite{nuscenes}~\textendash~Seq.~0664}}  &
    \includegraphics[clip=false, trim={0 0 0 0},width= \fgsize\columnwidth]{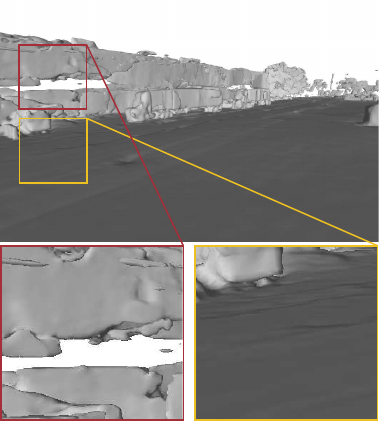} & 
    \includegraphics[clip=false, trim={0 0 0 0},width= \fgsize\columnwidth]{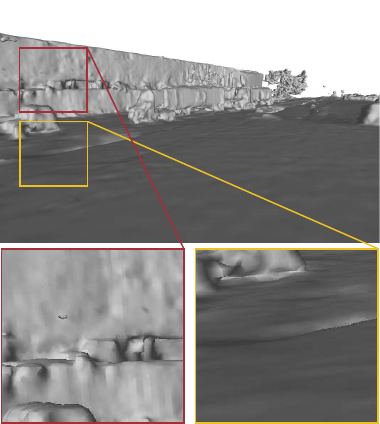} & 
    \includegraphics[clip=false, trim={0 0 0 0},width= \fgsize\columnwidth]{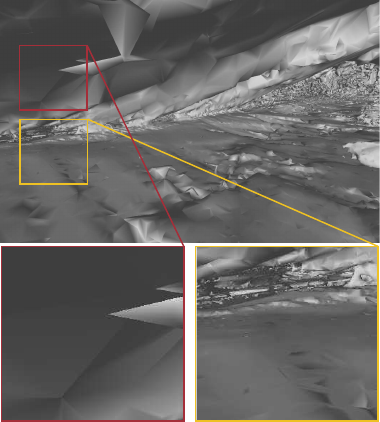} & 
    \includegraphics[clip=false, trim={0 0 0 0},width= \fgsize\columnwidth]{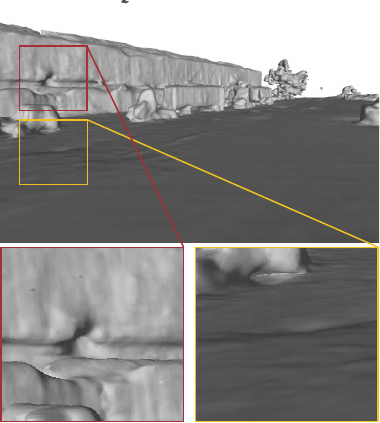} \\

\end{tabular}

  \caption{Qualitative experiments results on four popular autonomous driving datasets.  
  Complex scene geometries reconstructed by the mesh are color highlighted. We compare our mesh to the ones from StreetSurf~\cite{streetsurf}, ViiNeuS~\cite{scilla} and GoF~\cite{Yu2024GOF}}
\label{fig:mesh_comparison}

\end{figure*}

\paragraph{Qualitative analysis.}
To complete our evaluation, we further present qualitative geometric results in Figure~\ref{fig:mesh_comparison} across different dataset with complex scenes geometry, including a mixture of large road surfaces and fine structures such as poles, traffic lights, and tree trunks. Both ViiNeuS~\cite{scilla} and StreetSurf~\cite{streetsurf} are able to reconstruct smooth complete surface but result in incomplete fine structure reconstruction. GoF~\cite{Yu2024GOF} produce hight quality on fine structures but fails large road plane reconstruction (see sequence 0064 of nuScenes dataset). Our results demonstrate that our method outperforms the other baselines in accurately reconstructing both surfaces and fine structures, under conditions of limited image overlap.

\paragraph{Efficiency}

\begin{table}[tb]
    
    \centering
    \resizebox{0.8\columnwidth}{!}{%
    \begin{tabular}{lccc}
        & mesh extraction time & train time & params. size   \\
        & (infer.~\&~post-process. min)&(min)&(MiB)\\
        \midrule
        GOF & $30$ & $30-60$ & $100-300$   \\
        StreetSurf & $\approx~1$ & $40$ & $92.59$  \\
        ViiNeuS &  $<0.5$ & $20$ & $27$ \\
        \midrule
        
        \method~(ours) & $<0.5$ & $35$ & $51$ \\

    \end{tabular}
    }
    \vspace{-0.3cm}
        \caption{
        \footnotesize Performance was evaluated on the KITTI-360 dataset using the same consumer-grade GPU with 24GB of VRAM.}
        \vspace{-0.4cm}
    \label{tab:performances}
\end{table}

Table~\ref{tab:performances} reports the computational performance of each method. \method achieves the fastest mesh extraction time—under 30s—whereas GoF requires approximately 30min. Moreover, our approach delivers the highest-quality meshes while maintaining moderate training time and memory usage, a critical combination for large-scale autonomous driving applications.

\subsection{Ablation study}
\label{sec:ablation}
\begin{table}[!h]
\centering
\small
\begin{tabular}{cccc}
\hline
                    & P$\rightarrow$M (all)       & Prec. (all)     & P$\rightarrow$M (pole)      \\ \hline
w/o GRS & 0.11          & 0.79          & 0.46          \\
w/o masked regu.    & 0.13          & 0.75          & 0.53          \\
\hline
Full model              & \textbf{0.09} & \textbf{0.83} & \textbf{0.21} \\ \hline
\end{tabular}

\caption{Ablation of our contributions on the KITTI-360 dataset.\label{tab:ablation}}

\end{table}

To have a clear understanding the contribution of each key component in our method, we conducted an ablation study on the KITTI-360 dataset. The results are presented in Table \ref{tab:ablation} and Figure~\ref{fig:ablation}. We observe that both GRS and Adaptive Relaxation improve overall geometry, with Adaptive Relaxation proving essential for complete fine structures reconstruction (reducing the P$\rightarrow$M metric from 0.53 to 0.21). These findings align with the motivations behind our method design.

\section{Conclusion}

In this work, we presented \textbf{\method}, a novel uncertainty estimation framework that effectively combines volumetric SDF representations for robust urban scene reconstruction under limited view overlap. By estimating both photometric and geometric uncertainty, we introduce Guided Ray Sampling to deploy each representation where it excels. To avoid over regularization we propose a novel robust kernel to adaptively relax geometric regularization for SDF. Extensive quantitative and qualitative evaluations show that \method~outperforms SoTA SDF and Gaussian splatting methods, delivering more accurate reconstructions on both large planar regions and fine structures under autonomous driving sensor setting.

\clearpage
\setcounter{page}{1}
\maketitlesupplementary

\section{Additional implementation details}
\begin{table}[b]
    \centering
    \begin{tabular}{lcc}
         Parameter & Value \\
         \midrule
         Table size &  $2^{19}$\\
         Finest resolution  & 2048 \\
         Coarsest resolution & 16 \\ 
         Number of level & 16\\ 
        \bottomrule
    \end{tabular}
    \caption{Hash Grid encoding parameters for both volumetric representation and SDF representation}
    \label{tab:hashgrid}
\end{table}

\paragraph{Model architecture.}

For both representations $\mathcal{F}_\Theta^v$ and $\mathcal{F}_\Theta^f$, we use a hash grid for positional embedding~\cite{mueller2022instant} and spherical harmonics for directional embedding~\cite{kerbl3Dgaussians}. Additionally, we adopt the same appearance embedding from NeRF-W~\cite{martin2021nerfw} to address illumination changes. Both representations utilize independent hash grids, trained separately during the training process, with identical parameters as summarized in Table~\ref{tab:hashgrid}. We also employ two additional density estimators (proposal networks), $ \mathcal{F}_\Theta^{p_{0}}$ and $ \mathcal{F}_\Theta^{p_{1}}$~\cite{mip-nerf-360}, which are supervised only by the weight estimated by $\mathcal{F}_\Theta^v$, but will be used to guide the sampling for $\mathcal{F}_\Theta^f$ during initialization. The sampling parameters used for training in our experiments are detailed in Table~\ref{tab:num_sampling}. In the refinement stage of the SDF, we replace the estimated densities provided by the proposal networks $ \mathcal{F}_\Theta^{p_i}, i \in \{0,1\}$ with PDF weights computed directly using $\mathcal{F}_\Theta^f$. Additionally, we slightly increase the number of samples for both coarse and fine sampling, setting them to 32 and 28, respectively, to improve the accuracy of density estimation during the refinement stage.

\paragraph{Optimization.} The losses we use in \method~to optimize the volumetric ($\mathcal{L}_{\text{v}}$) and SDF fields ($\mathcal{L}_{\text{f}}$) are:
{\small
\begin{align}
    \mathcal{L}_{\text{v}} = 
    \mathcal{L}_{\text{rgb}} 
    + \lambda_d \mathcal{L}_d 
    + \lambda_{\text{sky}} \mathcal{L}_{\text{sky}} 
    + \lambda_{\hat{N}} \mathcal{L}_{\hat{N}}
    + \lambda_{\text{sem}} \mathcal{L}_{\text{sem}}, \\
    \mathcal{L}_{\text{f}} = 
    \mathcal{L}_{\text{rgb}} 
    + \lambda_d \mathcal{L}_d 
    + \lambda_{\text{sky}} \mathcal{L}_{\text{sky}} 
    + \lambda_{\hat{N}} \mathcal{L}_{\hat{N}}^{u}
    + \lambda_{\text{eik}} \mathcal{L}_{\text{eik}}^{u}.
\end{align}
}

We set $\lambda_{\text{sky}} = 0.01$, $\lambda_{d} = 0.001$, $\lambda_{\hat{N}} = 0.01$, $\lambda_{\text{sem}}=0.001$ and $\lambda_{\text{eik}} = 0.1$. During the SDF refinement stage, we increase $\lambda_{\hat{N}}$ to 0.05 only on flat semantic classes and $\lambda_d$ to 0.1. For Pandaset, we reduce the Eikonal loss coefficient to 0.01 during the early stage to facilitate mesh reconstruction, and then restore it to 0.1 for the final two epochs. To improve initialization on large texture-less road planes in the volumetric representation, we activate the total variation regularization on depth prediction during the first epoch for sequences 0034 and 0664 in nuScenes, and sequence 023 in Pandaset, with a weight of 1e-4.
\begin{table}[t]
    \centering
    \begin{tabular}{lcc}
         Model & Num. of Sampling \\
         \midrule
         $ \mathcal{F}_\Theta^{p_{0}}$ &  128\\
         $ \mathcal{F}_\Theta^{p_{1}}$ &  96 \\
         $\mathcal{F}_\Theta^v$ & 48 \\ 
         $\mathcal{F}_\Theta^f$ & 24 \\ 
        \bottomrule
    \end{tabular}
    \caption{Sampling parameters at Training}
    \label{tab:num_sampling}
\end{table}

\paragraph{Datasets.} For all experiments, we use the poses provided by the datasets, except Waymo, where inaccuracies in the data required recalculating the vehicle trajectory and sensor calibration using MOISST~\cite{Herau_2023}. The KITTI-360~\cite{Kitti} sequence splits used in our evaluations are summarized in Table~\ref{tab:kitti_seq}. For KITTI-360~\cite{Kitti}, we use all four cameras, while for Pandaset~\cite{pandaset}, nuScenes~\cite{nuscenes}, and Waymo Open Dataset~\cite{Waymo}, we use the three front-facing cameras. Additionally, we sample one image out of every two for KITTI-360 and one image out of every eight for the other datasets.
\begin{table}[b]
    \centering
    \small
    \begin{tabular}{lcccccccc}
         Seq. & KITTI Sync. &  Start & End  & \# frames per cam.\\
         \midrule
         30 &  0004 & 1728 & 1822 & 48\\
         31 & 0009 & 2890 & 2996 & 54\\
         35 & 0009 & 980 & 1092 & 57\\
        36 & 0010 & 112 & 166 & 28\\
        \bottomrule
    \end{tabular}
    \caption{Selected KITTI-360 sequences}
    \label{tab:kitti_seq}
\end{table}

\section{Additional details on Adaptive Thresholding}

\begin{figure}
\centering
\includegraphics[width=0.4\textwidth]{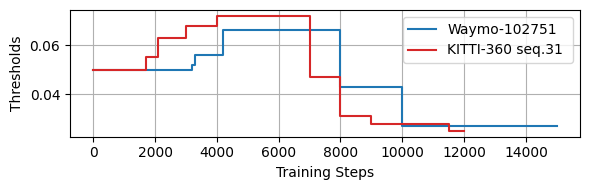} 
\vspace{-0.5cm}
\caption{ The similar convergence patterns of our adaptive threshold ($\tau$) on  KITTI-360 and Waymo datasets demonstrate it's generalization capability across diverse scenes. 
}
\label{fig:adaptive_thr}
\end{figure}

Figure~\ref{fig:adaptive_thr} complements the pseudocode in Algorithm 1 by illustrating how the uncertainty threshold dynamically adjusts during training. Notably, the threshold exhibits remarkably similar convergence patterns across different datasets (KITTI-360 and Waymo). This consistency demonstrates the strong generalization capability of our self-adaptive strategy across diverse scenes.

\section{Additional details on Adaptive Relaxation}

One of our key contributions is to use $\mathbb{I}_{\mu_c}$ to adaptively relax the Eikonal constrain and normal supervision on high frequency details and fine structures at early training stages (see Sections. 3.3).
Experimentally, we set the threshold  $\tau_c$  to 0.02 for KITTI and nuScenes, and to 0.015 for Pandaset and Waymo, as the images in these datasets are less contrastive.
\cref{fig:eik_mask} shows an visualization of how empirically $\mathbb{I}_{\mu_c}$ helps to divide the scene and how it evolves during training. 
In the initial stage, $\mathbb{I}_{\mu_c}$ indicates regions with strong "surfaceness", such as roads and buildings. By the end of the initialization stage, most surface regions are identified by  $\mathbb{I}_{\mu_c}$ and constrained by the Eikonal loss. During the refinement stage, as the SDF learns to capture fine structures and predict consistent mesh colors, the Eikonal loss is progressively applied to these regions as well, ensuring adherence to the properties of a signed distance function.
\section{Additional experiment results}

\begin{figure}
\centering
\begin{subfigure}{0.49\linewidth}
    \centering
    \includegraphics[width=1.0\textwidth]{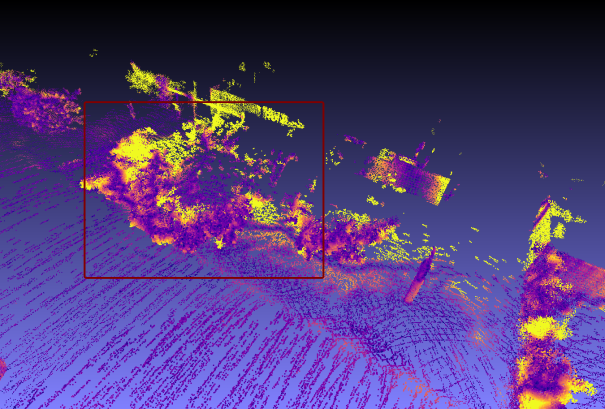} 
    \caption{StreetSurf}
\label{fig:occlusion_streetsurf}
\end{subfigure}
\begin{subfigure}{0.49\linewidth}
    \centering
    \includegraphics[width=1.0\textwidth]{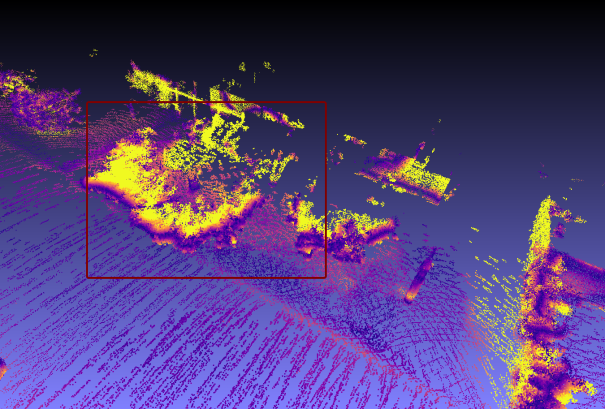} 
    \caption{\method~(ours)} 
    \label{fig:occlusion_ours}
\end{subfigure}\\

\vspace{-2mm}
\caption{Point-cloud colored by P$\rightarrow$M error in a heavily occluded scenario in Seq. 13196 from Waymo~\cite{Waymo}. Red boxes highlight densely occluded trees.}
\label{fig:occlusion}
\vspace{-5mm}
\end{figure}

\begin{figure*}[tb] 

\centering

\def\fgsize{0.48}
\def\rowspacing{0.2cm}

\scriptsize
\setlength{\tabcolsep}{0.0035\linewidth}
\renewcommand{\arraystretch}{1.0}
\begin{tabular}{ccccc}%

      & StreetSurf~\cite{streetsurf} & ViiNeuS~\cite{scilla} & GoF~\cite{Yu2024GOF}-dense& \method~(ours)\\

 \vspace{\rowspacing}
    \multirow{1}{*}[22mm]{\rotatebox[origin=c]{90}{KITTI-360~\cite{Kitti}~\textendash~Seq.~30}}  &
    \includegraphics[clip=false, trim={0 0 0 0},width= \fgsize\columnwidth]{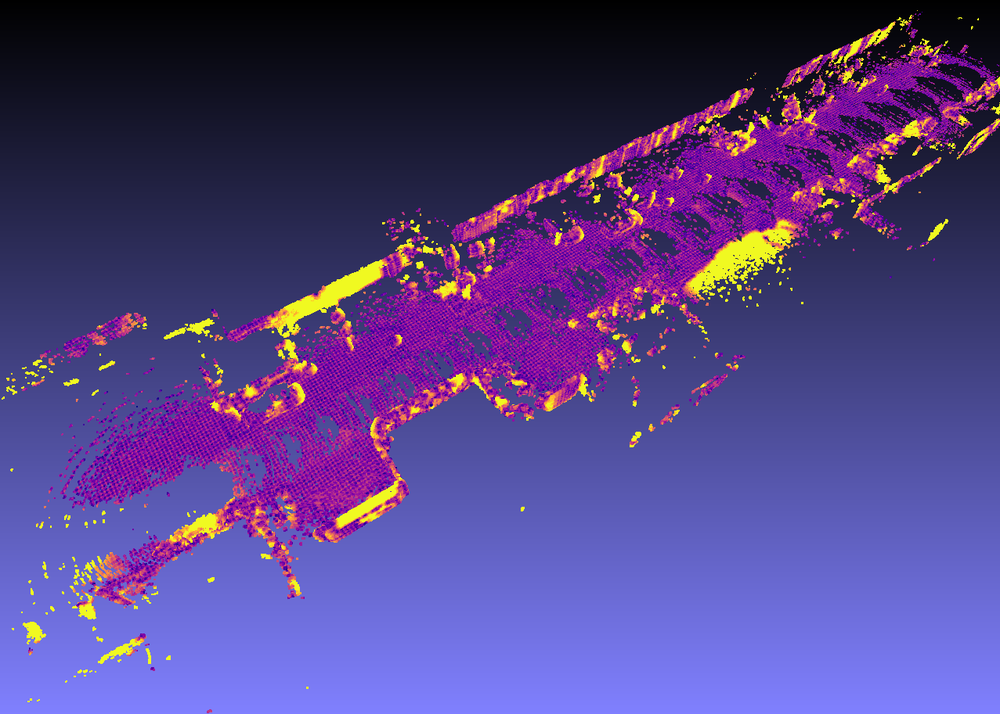} & 
    \includegraphics[clip=false, trim={0 0 0 0},width= \fgsize\columnwidth]{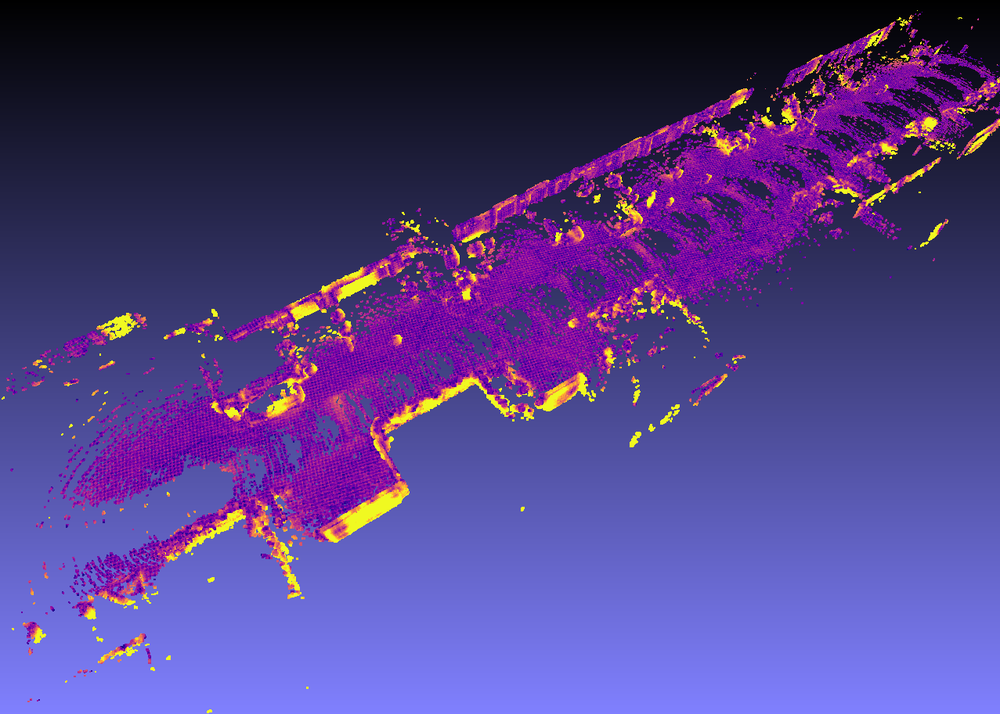} & 
    \includegraphics[clip=false, trim={0 0 0 0},width= \fgsize\columnwidth]{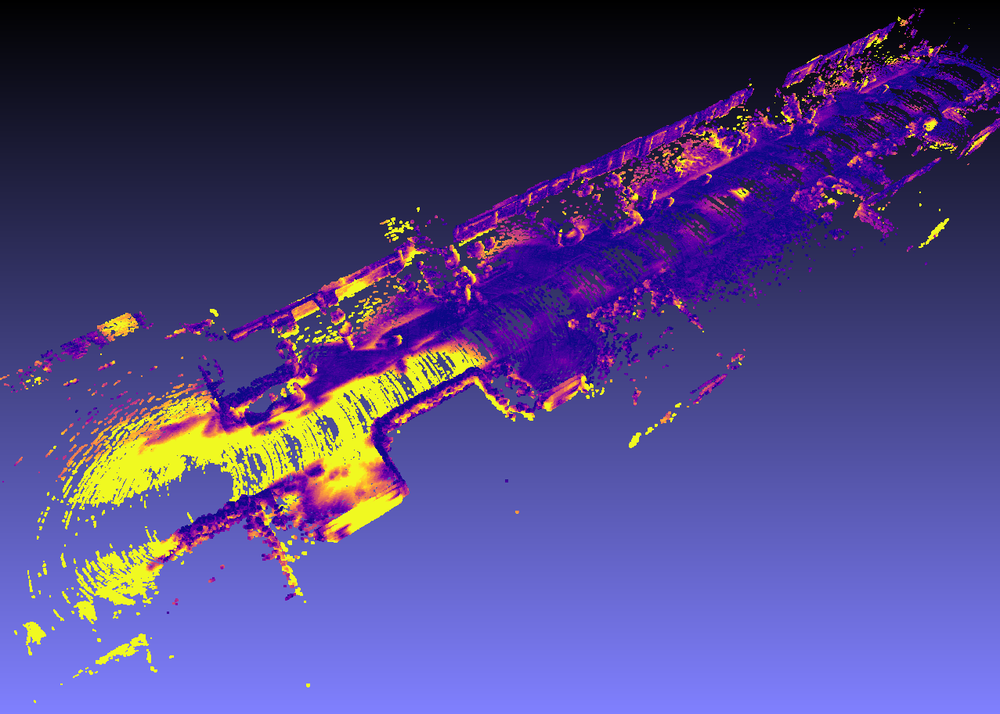} & 
    \includegraphics[clip=false, trim={0 0 0 0},width= \fgsize\columnwidth]{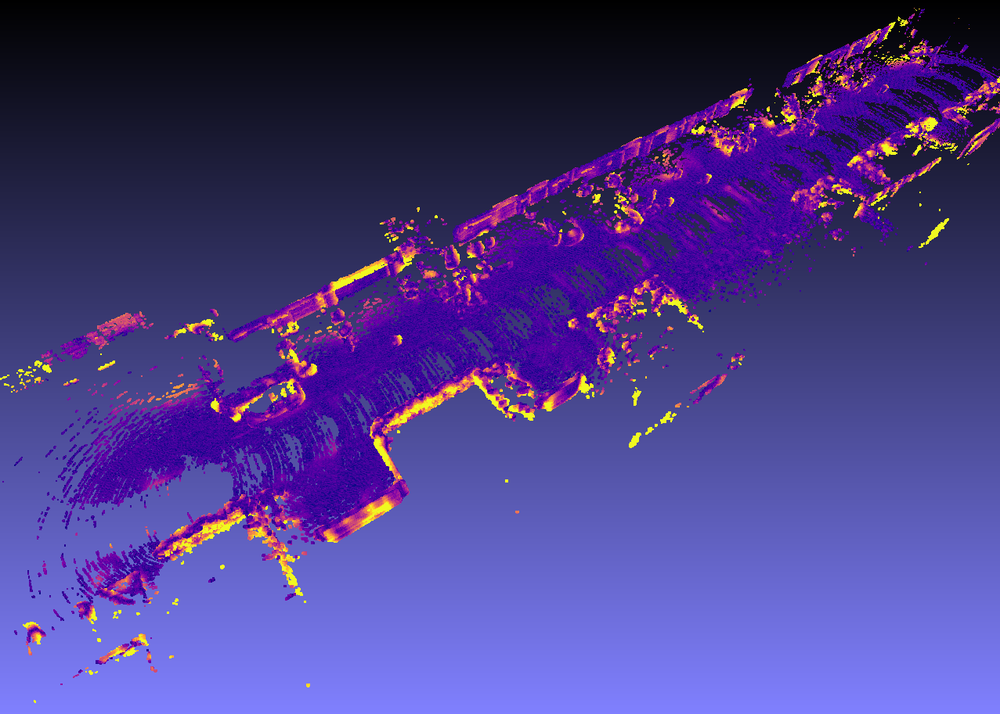} \\

    \vspace{\rowspacing}

    \multirow{1}{*}[24mm]{\rotatebox[origin=c]{90}{Pandaset~\cite{pandaset}~\textendash~Seq.~037}}  &
   \includegraphics[clip=false, trim={0 0 0 0},width= \fgsize\columnwidth]{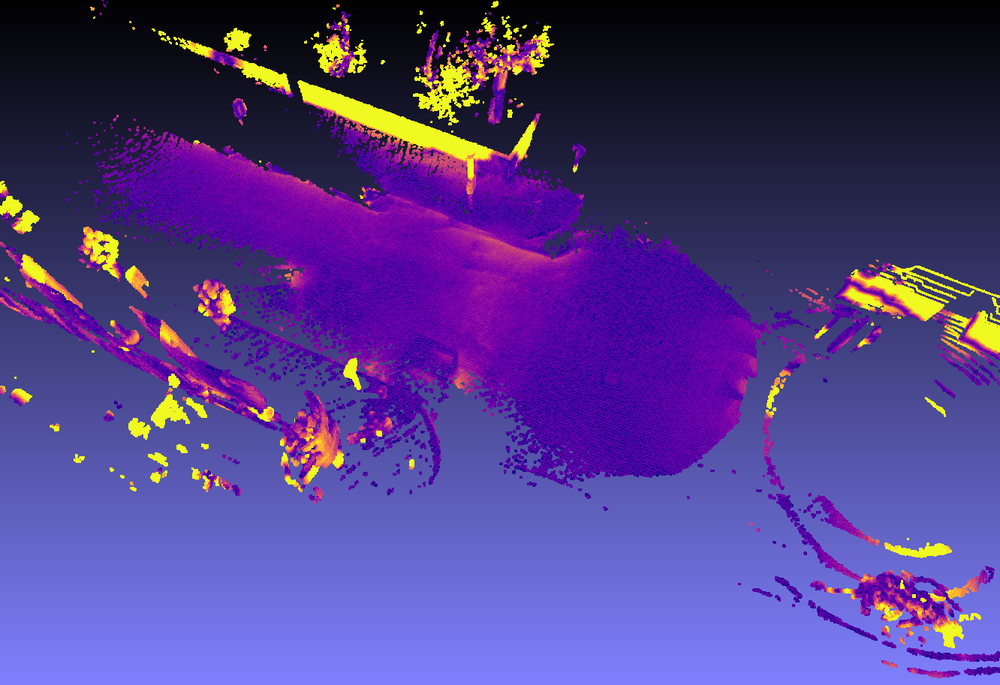} & 
    \includegraphics[clip=false, trim={0 0 0 0},width= \fgsize\columnwidth]{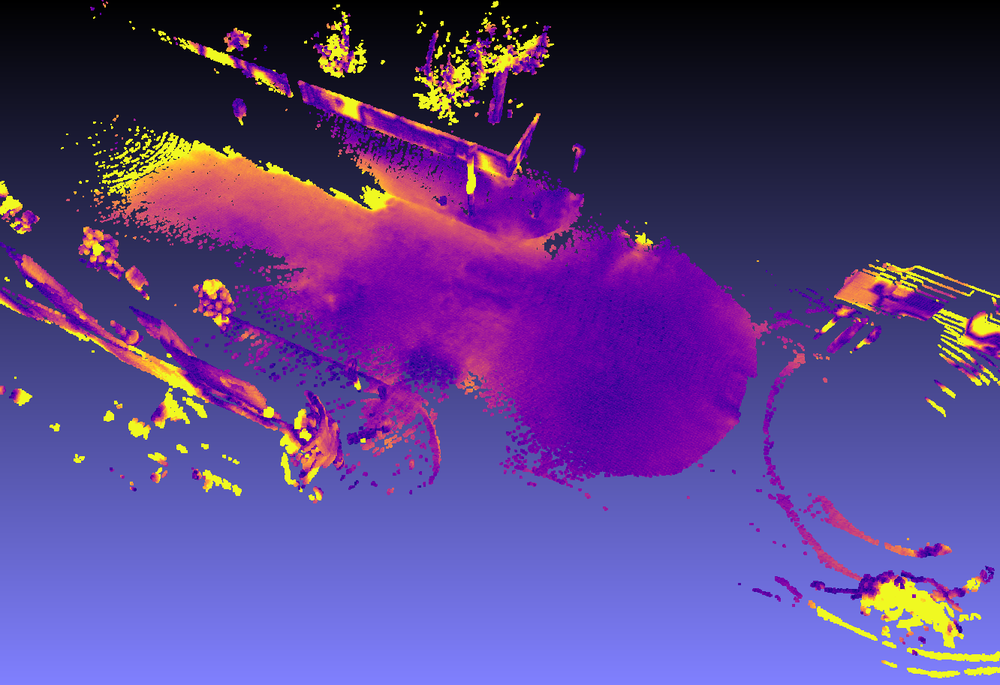} & 
    \includegraphics[clip=false, trim={0 0 0 0},width= \fgsize\columnwidth]{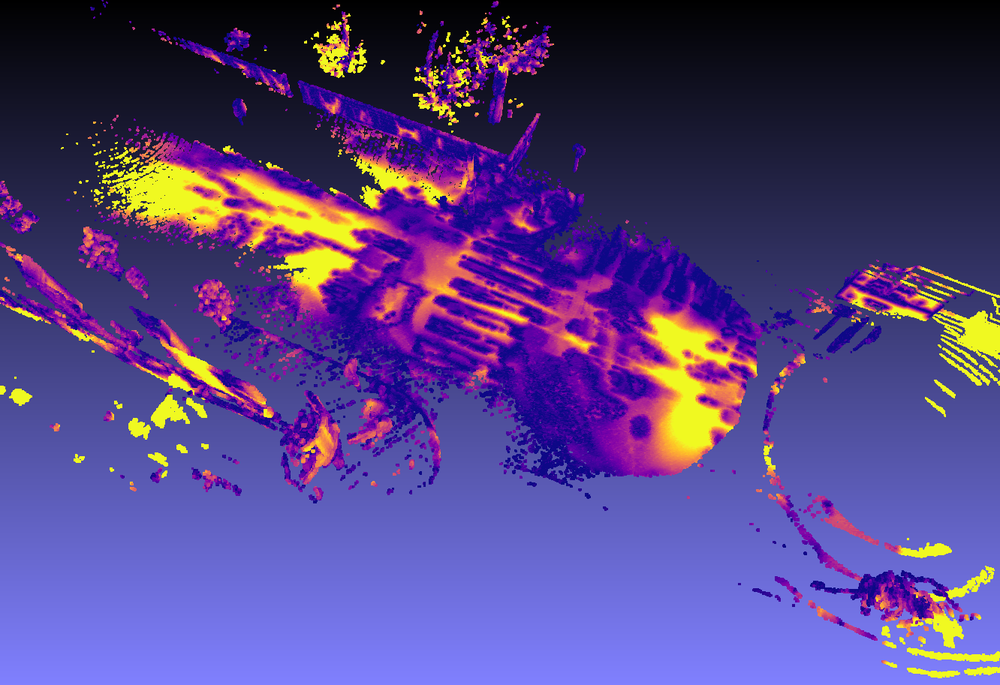} & 
    \includegraphics[clip=false, trim={0 0 0 0},width= \fgsize\columnwidth]{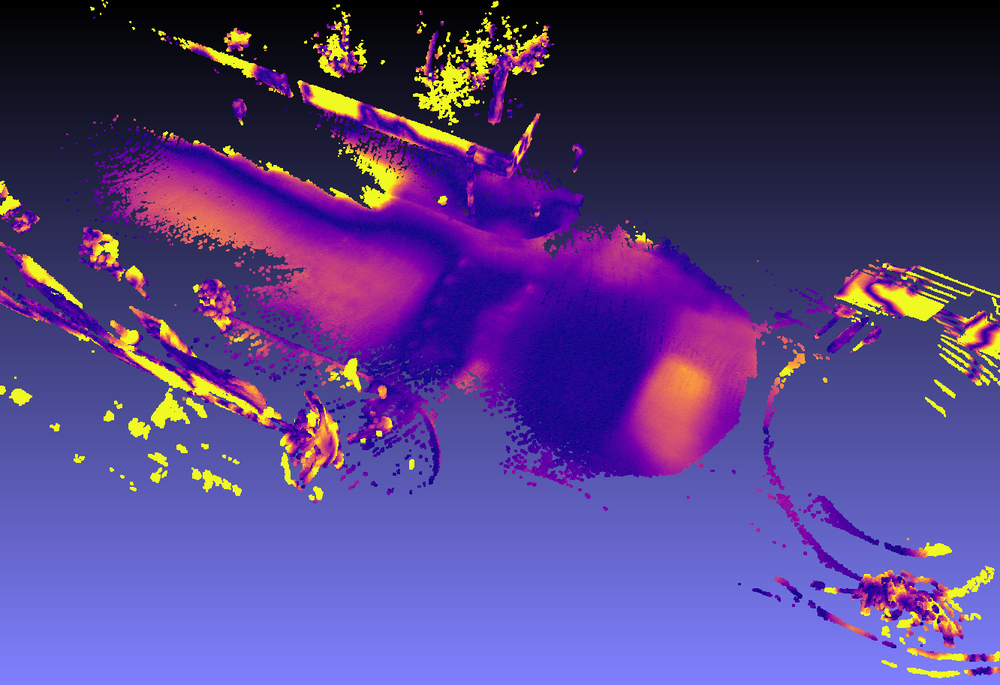} \\

 \vspace{\rowspacing}

    \multirow{1}{*}[25mm]{\rotatebox[origin=c]{90}{Waymo~\cite{Waymo}~\textendash~Seq.~14689}}  &
    \includegraphics[clip=false, trim={0 0 0 0},width= \fgsize\columnwidth]{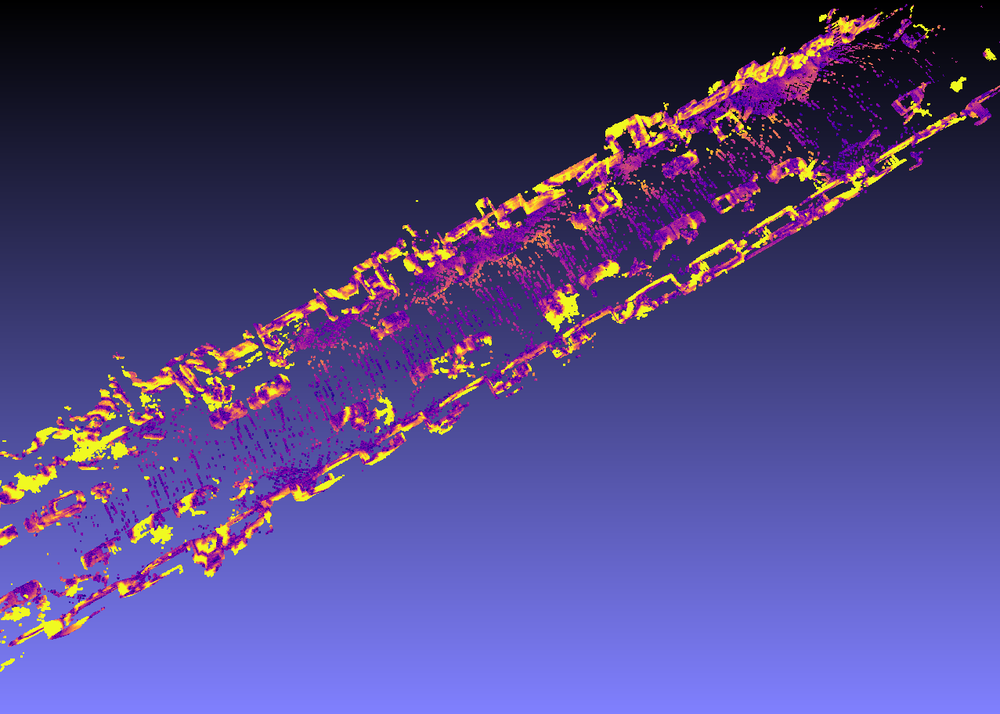} & 
    \includegraphics[clip=false, trim={0 0 0 0},width= \fgsize\columnwidth]{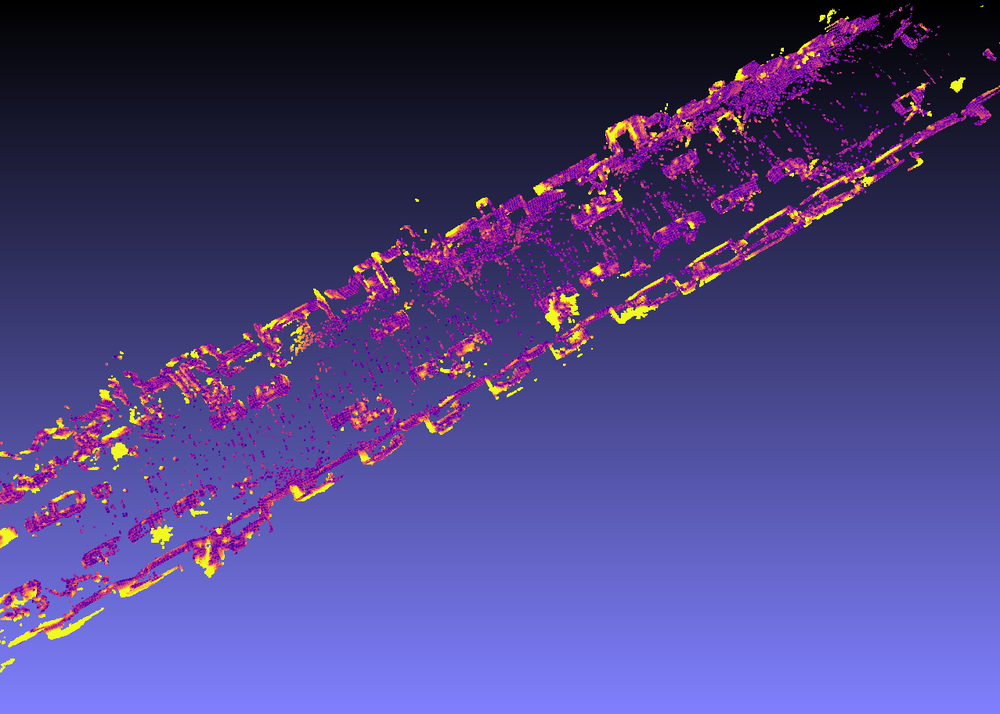} & 
    \includegraphics[clip=false, trim={0 0 0 0},width= \fgsize\columnwidth]{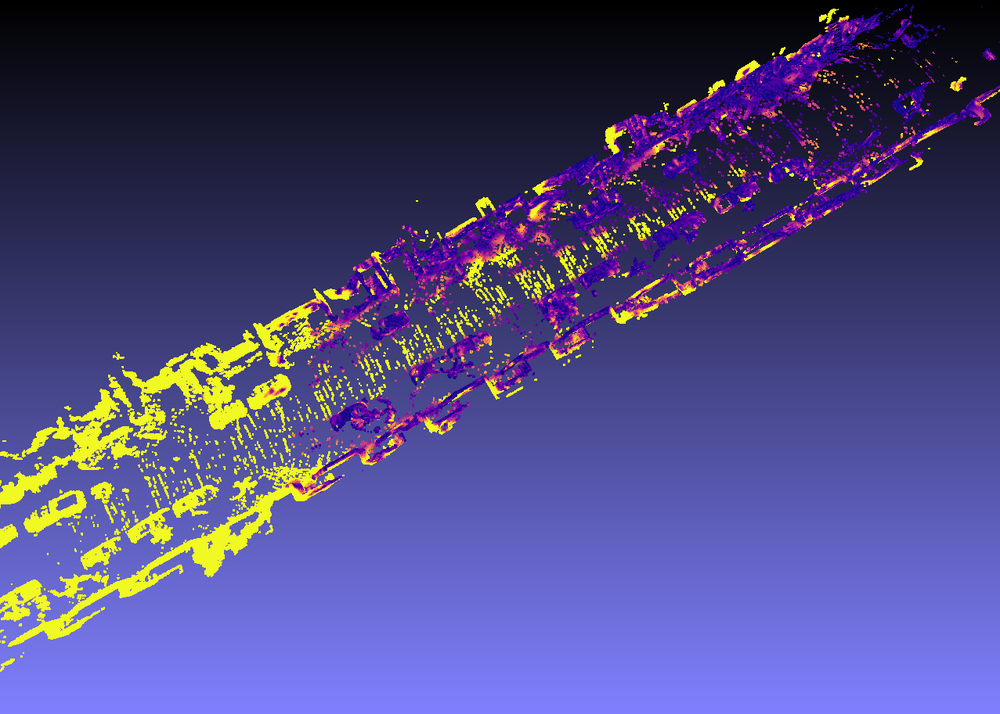} & 
    \includegraphics[clip=false, trim={0 0 0 0},width= \fgsize\columnwidth]{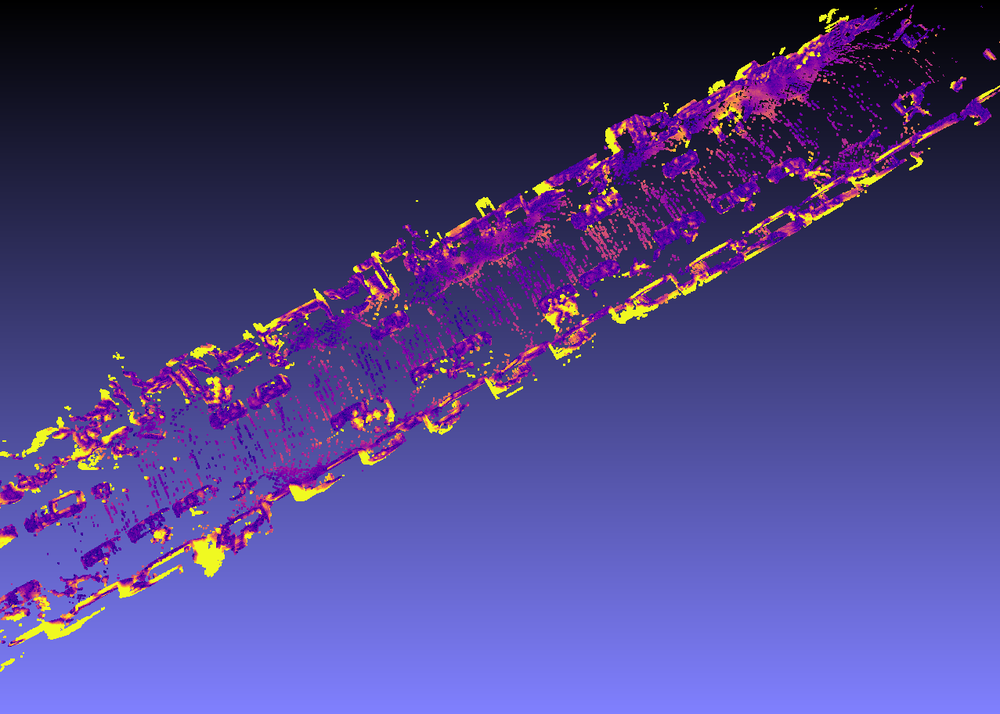} \\

     \vspace{\rowspacing}

    \multirow{1}{*}[25mm]{\rotatebox[origin=c]{90}{nuScenes~\cite{nuscenes}~\textendash~Seq.~0664}}  &
    \includegraphics[clip=false, trim={0 0 0 0},width= \fgsize\columnwidth]{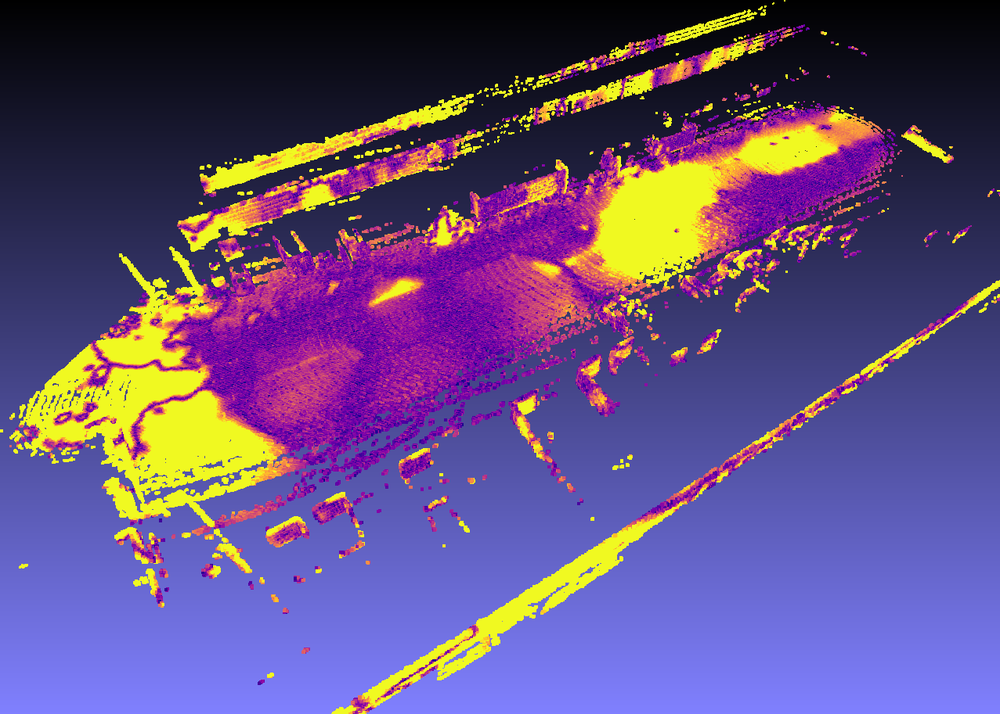} & 
    \includegraphics[clip=false, trim={0 0 0 0},width= \fgsize\columnwidth]{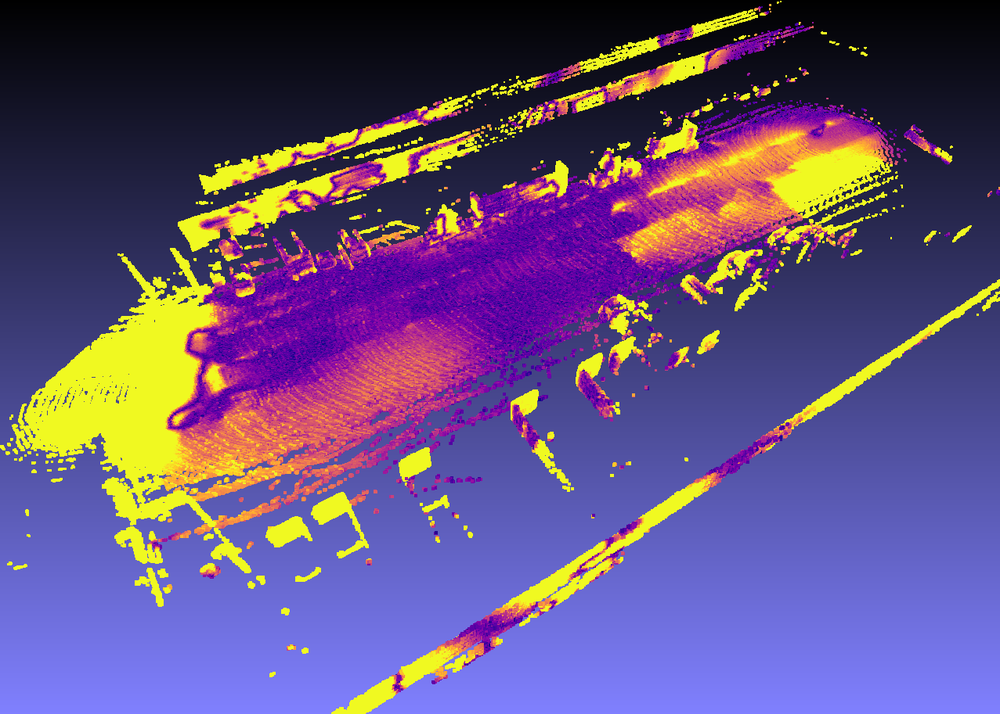} & 
    \includegraphics[clip=false, trim={0 0 0 0},width= \fgsize\columnwidth]{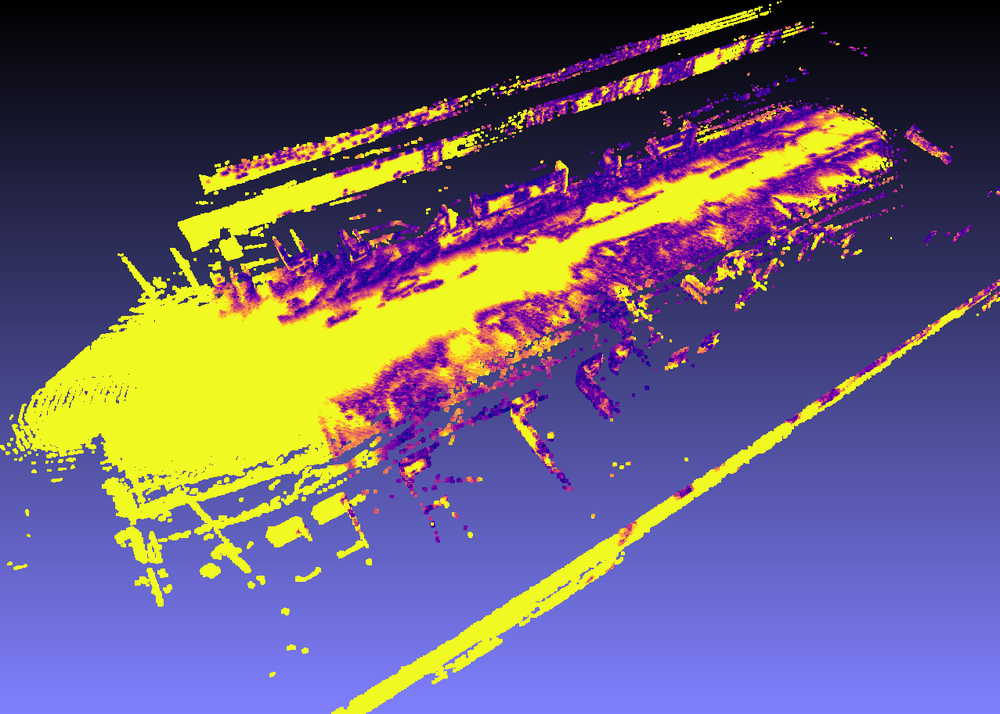} & 
    \includegraphics[clip=false, trim={0 0 0 0},width= \fgsize\columnwidth]{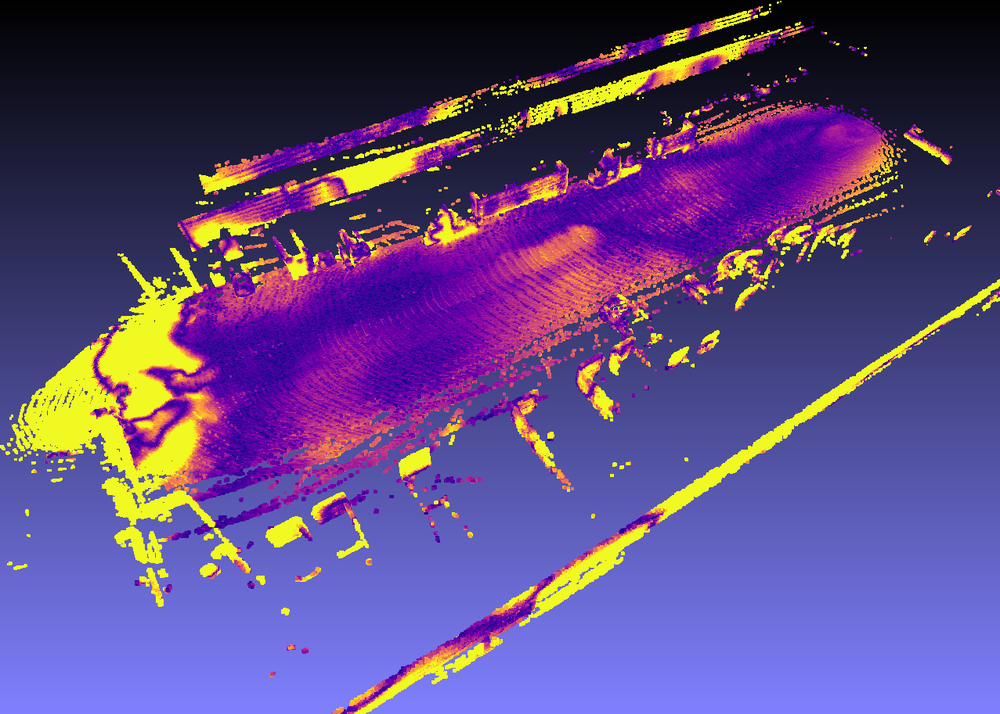} \\

    \multicolumn{5}{c}{\includegraphics[width=2.00\columnwidth]{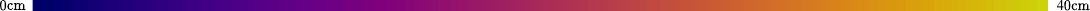}}

\end{tabular}

  \caption{Point-cloud colored by P$\rightarrow$M error on sequences reported in Figure 4 of the main paper. We compare our results to the ones from StreetSurf~\cite{streetsurf}, ViiNeuS~\cite{scilla} and GoF~\cite{Yu2024GOF}}
\label{fig:error_map}

\end{figure*}

\begin{figure*}[tb] 

\centering

\def\fgsize{0.48}
\def\rowspacing{0.2cm}

\scriptsize
\setlength{\tabcolsep}{0.0035\linewidth}
\renewcommand{\arraystretch}{1.0}
\begin{tabular}{ccccc}

      & StreetSurf~\cite{streetsurf} & ViiNeuS~\cite{scilla} & GoF~\cite{Yu2024GOF}-dense & \method~(ours)\\

 \vspace{\rowspacing}
    \multirow{1}{*}[31mm]{\rotatebox[origin=c]{90}{KITTI-360~\cite{Kitti}~\textendash~Seq.~35}}  &
    \includegraphics[clip=false, trim={0 0 0 0},width= \fgsize\columnwidth]{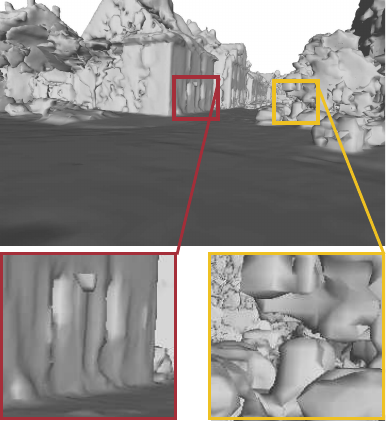} & 
    \includegraphics[clip=false, trim={0 0 0 0},width= \fgsize\columnwidth]{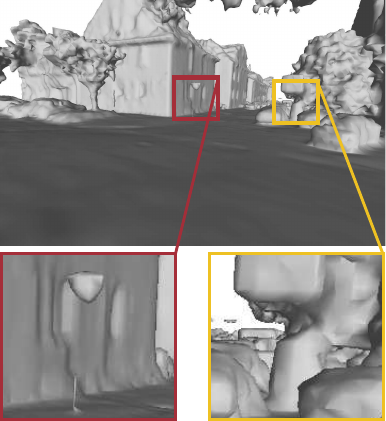} & 
    \includegraphics[clip=false, trim={0 0 0 0},width= \fgsize\columnwidth]{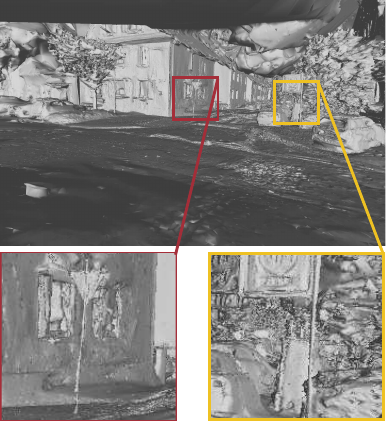} & 
    \includegraphics[clip=false, trim={0 0 0 0},width= \fgsize\columnwidth]{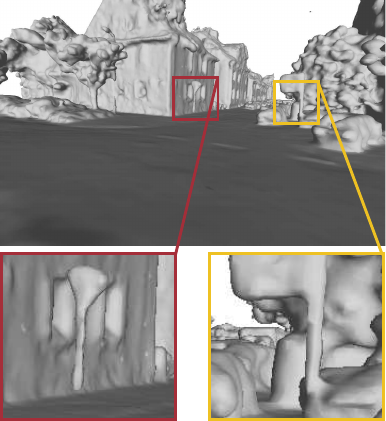} \\

    \vspace{\rowspacing}

    \multirow{1}{*}[33mm]{\rotatebox[origin=c]{90}{Pandaset~\cite{pandaset}~\textendash~Seq.~023}}  &
    \includegraphics[clip=false, trim={0 0 0 0},width= \fgsize\columnwidth]{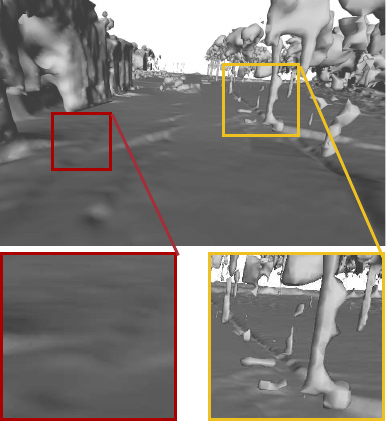} & 
    \includegraphics[clip=false, trim={0 0 0 0},width= \fgsize\columnwidth]{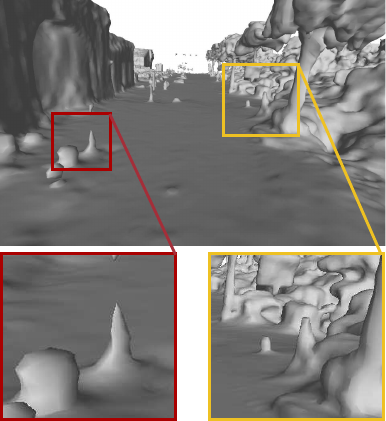} & 
    \includegraphics[clip=false, trim={0 0 0 0},width= \fgsize\columnwidth]{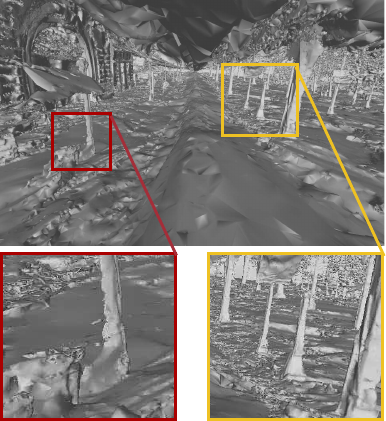} & 
    \includegraphics[clip=false, trim={0 0 0 0},width= \fgsize\columnwidth]{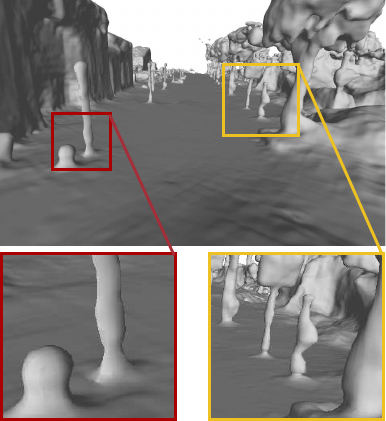} \\

 \vspace{\rowspacing}

    \multirow{1}{*}[34mm]{\rotatebox[origin=c]{90}{Waymo~\cite{Waymo}~\textendash~Seq.~102751}}  &
    \includegraphics[clip=false, trim={0 0 0 0},width= \fgsize\columnwidth]{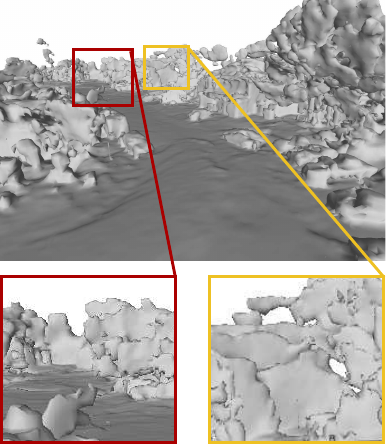} & 
    \includegraphics[clip=false, trim={0 0 0 0},width= \fgsize\columnwidth]{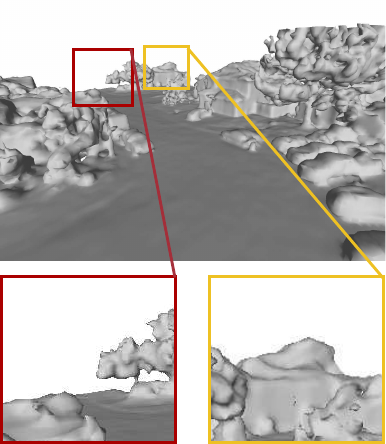} & 
    \includegraphics[clip=false, trim={0 0 0 0},width= \fgsize\columnwidth]{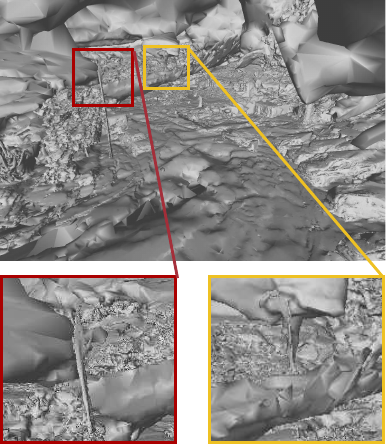} & 
    \includegraphics[clip=false, trim={0 0 0 0},width= \fgsize\columnwidth]{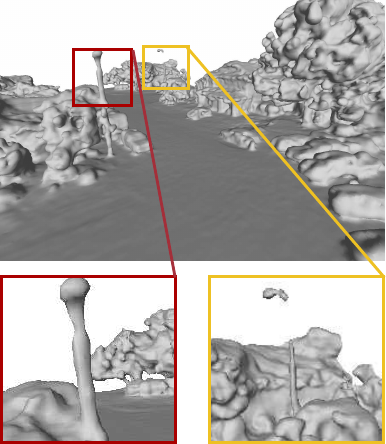} \\

     \vspace{\rowspacing}

    \multirow{1}{*}[33mm]{\rotatebox[origin=c]{90}{nuScenes~\cite{nuscenes}~\textendash~Seq.~0916}}  &
    \includegraphics[clip=false, trim={0 0 0 0},width= \fgsize\columnwidth]{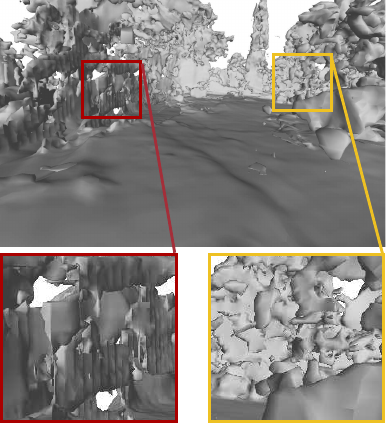} & 
    \includegraphics[clip=false, trim={0 0 0 0},width= \fgsize\columnwidth]{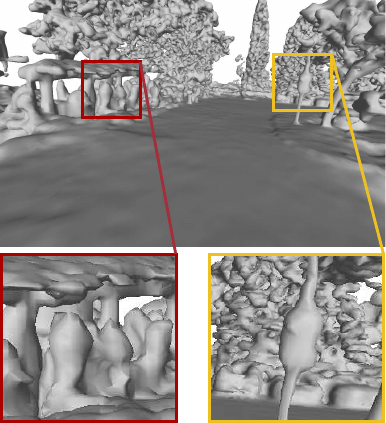} & 
    \includegraphics[clip=false, trim={0 0 0 0},width= \fgsize\columnwidth]{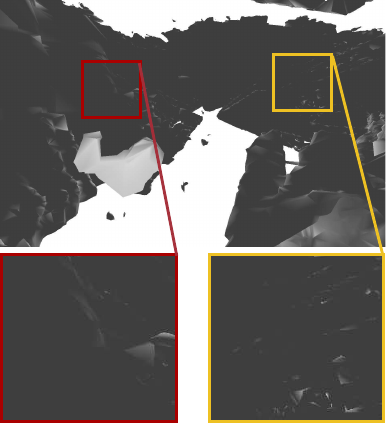} & 
    \includegraphics[clip=false, trim={0 0 0 0},width= \fgsize\columnwidth]{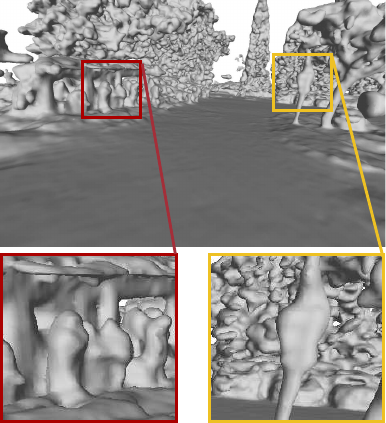} \\

\end{tabular}

  \caption{Additional qualitative results highlighting complex geometries. We compare our mesh to the ones generated from StreetSurf~\cite{streetsurf}, ViiNeuS~\cite{scilla} and GoF~\cite{Yu2024GOF}}
\label{fig:additional_vis_res}

\end{figure*}

\begin{figure*}[tb] 

\centering

\def\fgsize{0.9}
\def\rowspacing{0.2cm}

\scriptsize
\setlength{\tabcolsep}{0.0035\linewidth}
\renewcommand{\arraystretch}{1.0}
\begin{tabular}{ccc}

      & $\mathbb{I}_{\mu_c}$ & $\hat{C}_{\mathcal{E}}$\\

 \vspace{\rowspacing}
    \multirow{1}{*}[12mm]{\rotatebox[origin=c]{90}{Epoch~00}}  &
    \includegraphics[clip=false, trim={0 0 0 0},width= \fgsize\columnwidth]{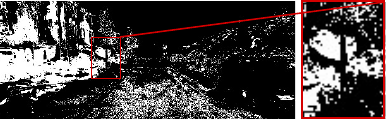} & 
    \includegraphics[clip=false, trim={0 0 0 0},width= \fgsize\columnwidth]{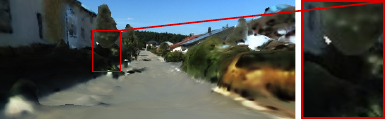}  \\

    \vspace{\rowspacing}

 \vspace{\rowspacing}
    \multirow{1}{*}[12mm]{\rotatebox[origin=c]{90}{Epoch~06}}  &
    \includegraphics[clip=false, trim={0 0 0 0},width= \fgsize\columnwidth]{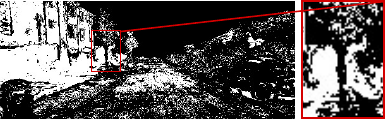} & 
    \includegraphics[clip=false, trim={0 0 0 0},width= \fgsize\columnwidth]{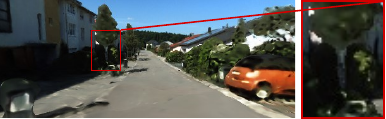}  \\

    \vspace{\rowspacing}

 \vspace{\rowspacing}
    \multirow{1}{*}[12mm]{\rotatebox[origin=c]{90}{Epoch~11}}  &
    \includegraphics[clip=false, trim={0 0 0 0},width= \fgsize\columnwidth]{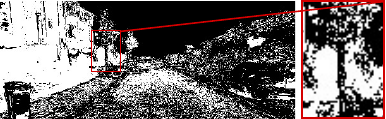} & 
    \includegraphics[clip=false, trim={0 0 0 0},width= \fgsize\columnwidth]{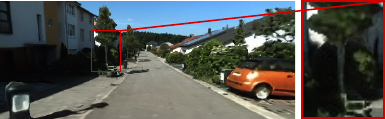}

\end{tabular}

  \caption{An example of how the robust kernel $\mathbb{I}_{\mu_c}$ evolves during the training stage on the KITTI-360 dataset. Black indicates region where $\mathcal{L}_{\text{eik}}$ and $\mathcal{L}_{\hat{N}}$ are relaxed for the SDF representation.}
\label{fig:eik_mask}

\end{figure*}

To provide a more comprehensive evaluation of the geometry throughout the scene, we include LiDAR point cloud visualizations colored by the P$\rightarrow$M distance in Figure~\ref{fig:error_map}. The results shown are the same sequences reported in Fig. 4 of the main paper. In the visualization, the color transitions from purple to yellow, indicating errors ranging from 0 cm to 40 cm. Overall, our method exhibits a larger proportion of purple regions and a reduced presence of yellow. Together, these results indicate that our approach achieves lower reconstruction errors across the entire scene, capturing both large surfaces and fine structures with higher accuracy.
Moreover, to provide a more thorough and detailed evaluation compared to other methods, we include additional qualitative comparisons in Figure~\ref{fig:additional_vis_res}. These visualizations further demonstrate the reconstruction quality of our approach, particularly in handling complex geometries and fine details.

In Figure~\ref{fig:occlusion}, we illustrate a case of densely occluded trees to further explain our suboptimal P$\rightarrow$M results in Waymo and Pandaset. Because GRS adjusts sampling to the nearest visible surface, it creates a “shell” that encloses occluded points, leading to lower P$\rightarrow$M metrics when compared to the dense LiDAR ground truth. 
In contrast, StreetSurf, which models mid-range objects using a density field, generates volumetric content within the tree region.

\section{Limitations}
Although our method results in overall high mesh quality on both large surfaces and fine structures, it still struggles in certain cases to deliver high-quality reconstructions. These challenges are particularly apparent with extremely fine and repetitive structures (\eg, iron railings), or when such structures are located in heavily occluded environments (\eg, areas surrounded by dense trees). Possible reasons include the inability of both representations to produce valid predictions in these scenarios and the guided sampling focusing on incorrect surfaces.

\begin{figure*}
\centering
\begin{subfigure}{0.49\linewidth}
    \centering
    \includegraphics[width=1.0\textwidth]{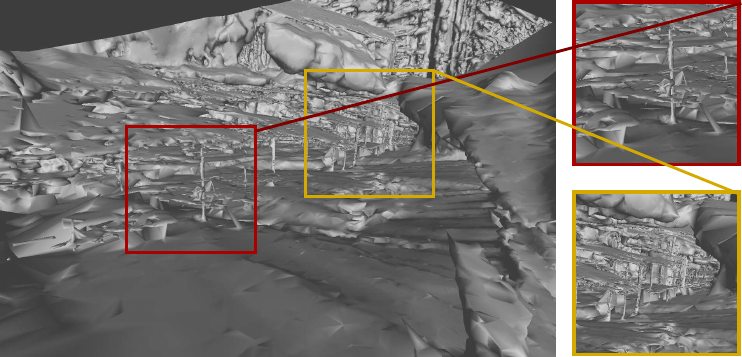} 
    \caption{GoF~\cite{Yu2024GOF}}
\end{subfigure}
\begin{subfigure}{0.49\linewidth}
    \centering
    \includegraphics[width=1.0\textwidth]{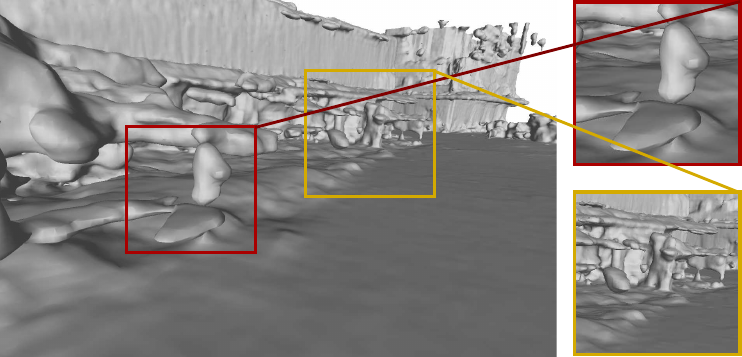} 
    \caption{\method~(ours)} 
\end{subfigure}\\
\begin{subfigure}{0.49\linewidth}
    \centering
    \includegraphics[width=1.0\textwidth]{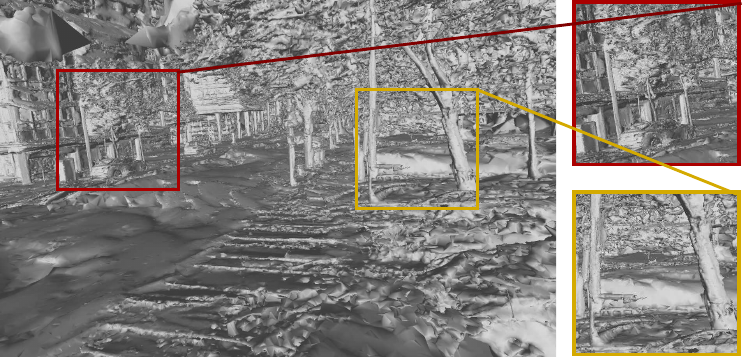} 
    \caption{GoF~\cite{Yu2024GOF}}
\end{subfigure}
\begin{subfigure}{0.49\linewidth}
    \centering
    \includegraphics[width=1.0\textwidth]{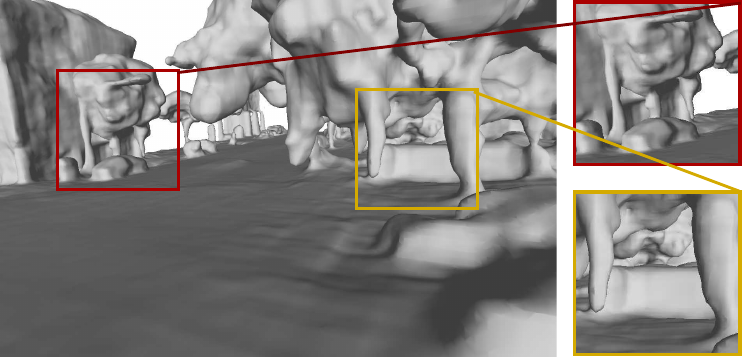} 
    \caption{\method~(ours)} 
\end{subfigure}\\

\caption{Examples of challenging scenarios. Iron railings from the nuScenes sequence~0072 (top) and dense vegetation occlusion from the Pandaset sequence~043 (bottom).\label{fig:failure_case_occ}\label{fig:failure_case_fine}}
\end{figure*}

Interestingly, since 3D Gaussians are a more flexible and less constrained representation compared to SDF, GoF~\cite{Yu2024GOF} performs better at capturing fine structures in such cases, even though it fails at large surface reconstruction under conditions of limited overlap. Applying our key concept of "joint" optimization with uncertainty estimation to representations like GoF could be a promising direction for future exploration. We illustrate such cases in Figure~\ref{fig:failure_case_fine}.

\clearpage

{
    \small
    \bibliographystyle{ieeenat_fullname}
    \bibliography{main}

\begin{thebibliography}{44}
\providecommand{\natexlab}[1]{#1}
\providecommand{\url}[1]{\texttt{#1}}
\expandafter\ifx\csname urlstyle\endcsname\relax
  \providecommand{\doi}[1]{doi: #1}\else
  \providecommand{\doi}{doi: \begingroup \urlstyle{rm}\Url}\fi

\bibitem[Barron et~al.(2022)Barron, Mildenhall, Verbin, Srinivasan, and Hedman]{mip-nerf-360}
Jonathan~T. Barron, Ben Mildenhall, Dor Verbin, Pratul~P. Srinivasan, and Peter Hedman.
\newblock Mip-nerf 360: Unbounded anti-aliased neural radiance fields.
\newblock In \emph{CVPR}, 2022.

\bibitem[Caesar et~al.(2020)Caesar, Bankiti, Lang, Vora, Liong, Xu, Krishnan, Pan, Baldan, and Beijbom]{nuscenes}
Holger Caesar, Varun Bankiti, Alex~H. Lang, Sourabh Vora, Venice~Erin Liong, Qiang Xu, Anush Krishnan, Yu Pan, Giancarlo Baldan, and Oscar Beijbom.
\newblock nuscenes: A multimodal dataset for autonomous driving.
\newblock In \emph{CVPR}, 2020.

\bibitem[Chen et~al.(2024)Chen, Li, Ye, Wang, Xie, Zhai, Wang, Liu, Bao, and Zhang]{chen2024pgsrplanarbasedgaussiansplatting}
Danpeng Chen, Hai Li, Weicai Ye, Yifan Wang, Weijian Xie, Shangjin Zhai, Nan Wang, Haomin Liu, Hujun Bao, and Guofeng Zhang.
\newblock Pgsr: Planar-based gaussian splatting for efficient and high-fidelity surface reconstruction, 2024.

\bibitem[Chen et~al.(2023)Chen, Li, and Lee]{chen2023neusg}
Hanlin Chen, Chen Li, and Gim~Hee Lee.
\newblock Neusg: Neural implicit surface reconstruction with 3d gaussian splatting guidance, 2023.

\bibitem[Chen et~al.(2022)Chen, Wang, Guo, and Zhang]{structnerf}
Zheng Chen, Chen Wang, Yuan-Chen Guo, and Song-Hai Zhang.
\newblock Structnerf: Neural radiance fields for indoor scenes with structural hints, 2022.

\bibitem[Cheng et~al.(2022)Cheng, Misra, Schwing, Kirillov, and Girdhar]{cheng2021mask2former}
Bowen Cheng, Ishan Misra, Alexander~G. Schwing, Alexander Kirillov, and Rohit Girdhar.
\newblock Masked-attention mask transformer for universal image segmentation.
\newblock In \emph{CVPR}, 2022.

\bibitem[Cui et~al.(2024)Cui, Ye, Wang, Zhang, Zhou, and Li]{streetsurfgs}
Xiao Cui, Weicai Ye, Yifan Wang, Guofeng Zhang, Wengang Zhou, and Houqiang Li.
\newblock Streetsurfgs: Scalable urban street surface reconstruction with planar-based gaussian splatting, 2024.

\bibitem[Deng et~al.(2022)Deng, Liu, Zhu, and Ramanan]{deng2022depth}
Kangle Deng, Andrew Liu, Jun-Yan Zhu, and Deva Ramanan.
\newblock Depth-supervised nerf: Fewer views and faster training for free.
\newblock In \emph{Proceedings of the IEEE/CVF Conference on Computer Vision and Pattern Recognition}, pages 12882--12891, 2022.

\bibitem[Djeghim et~al.(2025)Djeghim, Piasco, Bennehar, Rold{\~a}o, Tsishkou, and Sidib{\'e}]{scilla}
Hala Djeghim, Nathan Piasco, Moussab Bennehar, Luis Rold{\~a}o, Dzmitry Tsishkou, and D{\'e}sir{\'e} Sidib{\'e}.
\newblock {ViiNeuS}: Volumetric initialization for implicit neural surface reconstruction of urban scenes with limited image overlap.
\newblock In \emph{Proceedings of the IEEE/CVF Conference on Computer Vision and Pattern Recognition}, 2025.

\bibitem[Eftekhar et~al.(2021)Eftekhar, Sax, Malik, and Zamir]{eftekhar2021omnidata}
Ainaz Eftekhar, Alexander Sax, Jitendra Malik, and Amir Zamir.
\newblock Omnidata: A scalable pipeline for making multi-task mid-level vision datasets from 3d scans.
\newblock In \emph{ICCV}, 2021.

\bibitem[Gu\'edon and Lepetit(2024)]{Guedon_2024_CVPR}
Antoine Gu\'edon and Vincent Lepetit.
\newblock Sugar: Surface-aligned gaussian splatting for efficient 3d mesh reconstruction and high-quality mesh rendering.
\newblock In \emph{CVPR}, 2024.

\bibitem[Guo et~al.(2023)Guo, Deng, Li, Bai, Shi, Wang, Ding, Wang, and Li]{streetsurf}
Jianfei Guo, Nianchen Deng, Xinyang Li, Yeqi Bai, Botian Shi, Chiyu Wang, Chenjing Ding, Dongliang Wang, and Yikang Li.
\newblock Streetsurf: Extending multi-view implicit surface reconstruction to street views, 2023.

\bibitem[Herau et~al.(2023)Herau, Piasco, Bennehar, Roldão, Tsishkou, Migniot, Vasseur, and Demonceaux]{Herau_2023}
Quentin Herau, Nathan Piasco, Moussab Bennehar, Luis Roldão, Dzmitry Tsishkou, Cyrille Migniot, Pascal Vasseur, and Cédric Demonceaux.
\newblock Moisst: Multimodal optimization of implicit scene for spatiotemporal calibration.
\newblock In \emph{IROS}. IEEE, 2023.

\bibitem[Huang et~al.(2024)Huang, Yu, Chen, Geiger, and Gao]{huang20242d}
Binbin Huang, Zehao Yu, Anpei Chen, Andreas Geiger, and Shenghua Gao.
\newblock 2d gaussian splatting for geometrically accurate radiance fields.
\newblock In \emph{ACM SIGGRAPH 2024 Conference Papers}, pages 1--11, 2024.

\bibitem[Kazhdan et~al.(2006)Kazhdan, Bolitho, and Hoppe]{kazhdan2006poisson}
Michael Kazhdan, Matthew Bolitho, and Hugues Hoppe.
\newblock Poisson surface reconstruction.
\newblock In \emph{Eurographics Symp. Geometry Processing}, page 61–70, 2006.

\bibitem[Kerbl et~al.(2023)Kerbl, Kopanas, Leimk{\"u}hler, and Drettakis]{kerbl3Dgaussians}
Bernhard Kerbl, Georgios Kopanas, Thomas Leimk{\"u}hler, and George Drettakis.
\newblock 3d gaussian splatting for real-time radiance field rendering.
\newblock \emph{ACM Transactions on Graphics}, 42\penalty0 (4), 2023.

\bibitem[Li et~al.(2023)Li, M\"uller, Evans, Taylor, Unberath, Liu, and Lin]{li2023neuralangelo}
Zhaoshuo Li, Thomas M\"uller, Alex Evans, Russell~H Taylor, Mathias Unberath, Ming-Yu Liu, and Chen-Hsuan Lin.
\newblock Neuralangelo: High-fidelity neural surface reconstruction.
\newblock In \emph{CVPR}, 2023.

\bibitem[Liao et~al.(2022)Liao, Xie, and Geiger]{Kitti}
Yiyi Liao, Jun Xie, and Andreas Geiger.
\newblock {KITTI}-360: A novel dataset and benchmarks for urban scene understanding in 2d and 3d.
\newblock \emph{PAMI}, 2022.

\bibitem[Lorensen and Cline(1998)]{lorensen1998marching}
William~E Lorensen and Harvey~E Cline.
\newblock Marching cubes: A high resolution 3d surface construction algorithm.
\newblock In \emph{Seminal graphics: pioneering efforts that shaped the field}, pages 347--353, 1998.

\bibitem[Martin-Brualla et~al.(2021)Martin-Brualla, Radwan, Sajjadi, Barron, Dosovitskiy, and Duckworth]{martin2021nerfw}
Ricardo Martin-Brualla, Noha Radwan, Mehdi~SM Sajjadi, Jonathan~T Barron, Alexey Dosovitskiy, and Daniel Duckworth.
\newblock Nerf in the wild: Neural radiance fields for unconstrained photo collections.
\newblock In \emph{Proceedings of the IEEE/CVF conference on computer vision and pattern recognition}, pages 7210--7219, 2021.

\bibitem[Mildenhall et~al.(2020)Mildenhall, Srinivasan, Tancik, Barron, Ramamoorthi, and Ng]{2020nerf}
Ben Mildenhall, Pratul~P. Srinivasan, Matthew Tancik, Jonathan~T. Barron, Ravi Ramamoorthi, and Ren Ng.
\newblock Nerf: Representing scenes as neural radiance fields for view synthesis.
\newblock In \emph{ECCV}, 2020.

\bibitem[Moulon et~al.(2016)Moulon, Monasse, Perrot, and Marlet]{openMVG}
Pierre Moulon, Pascal Monasse, Romuald Perrot, and Renaud Marlet.
\newblock Open{MVG}: Open multiple view geometry.
\newblock In \emph{International Workshop on Reproducible Research in Pattern Recognition}, pages 60--74. Springer, 2016.

\bibitem[M\"uller et~al.(2022)M\"uller, Evans, Schied, and Keller]{mueller2022instant}
Thomas M\"uller, Alex Evans, Christoph Schied, and Alexander Keller.
\newblock Instant neural graphics primitives with a multiresolution hash encoding.
\newblock \emph{ACM Trans. Graph.}, 41\penalty0 (4):\penalty0 102:1--102:15, 2022.

\bibitem[Niemeyer et~al.(2022)Niemeyer, Barron, Mildenhall, Sajjadi, Geiger, and Radwan]{niemeyer2022regnerf}
Michael Niemeyer, Jonathan~T Barron, Ben Mildenhall, Mehdi~SM Sajjadi, Andreas Geiger, and Noha Radwan.
\newblock Regnerf: Regularizing neural radiance fields for view synthesis from sparse inputs.
\newblock In \emph{Proceedings of the IEEE/CVF Conference on Computer Vision and Pattern Recognition}, pages 5480--5490, 2022.

\bibitem[Pun et~al.(2023)Pun, Sun, Wang, Chen, Yang, Manivasagam, Ma, and Urtasun]{Lightsim}
Ava Pun, Gary Sun, Jingkang Wang, Yun Chen, Ze Yang, Sivabalan Manivasagam, Wei-Chiu Ma, and Raquel Urtasun.
\newblock Lightsim: Neural lighting simulation for urban scenes, 2023.

\bibitem[Rematas et~al.(2022{\natexlab{a}})Rematas, Liu, Srinivasan, Barron, Tagliasacchi, Funkhouser, and Ferrari]{rematas2022urban}
Konstantinos Rematas, Andrew Liu, Pratul~P Srinivasan, Jonathan~T Barron, Andrea Tagliasacchi, Thomas Funkhouser, and Vittorio Ferrari.
\newblock Urban radiance fields.
\newblock In \emph{Proceedings of the IEEE/CVF Conference on Computer Vision and Pattern Recognition}, pages 12932--12942, 2022{\natexlab{a}}.

\bibitem[Rematas et~al.(2022{\natexlab{b}})Rematas, Liu, Srinivasan, Barron, Tagliasacchi, Funkhouser, and Ferrari]{urban-radiance-fields}
Konstantinos Rematas, Andrew Liu, Pratul~P. Srinivasan, Jonathan~T. Barron, Andrea Tagliasacchi, Tom Funkhouser, and Vittorio Ferrari.
\newblock Urban radiance fields.
\newblock In \emph{CVPR}, 2022{\natexlab{b}}.

\bibitem[Sabour et~al.(2023)Sabour, Vora, Duckworth, Krasin, Fleet, and Tagliasacchi]{sabour2023robustnerf}
Sara Sabour, Suhani Vora, Daniel Duckworth, Ivan Krasin, David~J Fleet, and Andrea Tagliasacchi.
\newblock Robustnerf: Ignoring distractors with robust losses.
\newblock In \emph{Proceedings of the IEEE/CVF Conference on Computer Vision and Pattern Recognition}, pages 20626--20636, 2023.

\bibitem[Sun et~al.(2020)Sun, Kretzschmar, Dotiwalla, Chouard, Patnaik, Tsui, Guo, Zhou, Chai, Caine, Vasudevan, Han, Ngiam, Zhao, Timofeev, Ettinger, Krivokon, Gao, Joshi, Zhang, Shlens, Chen, and Anguelov]{Waymo}
Pei Sun, Henrik Kretzschmar, Xerxes Dotiwalla, Aurelien Chouard, Vijaysai Patnaik, Paul Tsui, James Guo, Yin Zhou, Yuning Chai, Benjamin Caine, Vijay Vasudevan, Wei Han, Jiquan Ngiam, Hang Zhao, Aleksei Timofeev, Scott Ettinger, Maxim Krivokon, Amy Gao, Aditya Joshi, Yu Zhang, Jonathon Shlens, Zhifeng Chen, and Dragomir Anguelov.
\newblock Scalability in perception for autonomous driving: Waymo open dataset.
\newblock In \emph{CVPR}, 2020.

\bibitem[Turki et~al.(2024)Turki, Agrawal, Bul{\`o}, Porzi, Kontschieder, Ramanan, Zollh{\"o}fer, and Richardt]{turki2024hybridnerf}
Haithem Turki, Vasu Agrawal, Samuel~Rota Bul{\`o}, Lorenzo Porzi, Peter Kontschieder, Deva Ramanan, Michael Zollh{\"o}fer, and Christian Richardt.
\newblock Hybridnerf: Efficient neural rendering via adaptive volumetric surfaces.
\newblock In \emph{Proceedings of the IEEE/CVF Conference on Computer Vision and Pattern Recognition}, pages 19647--19656, 2024.

\bibitem[Wang et~al.(2024{\natexlab{a}})Wang, Louys, Piasco, Bennehar, Roldão, and Tsishkou]{wang2023planerf}
Fusang Wang, Arnaud Louys, Nathan Piasco, Moussab Bennehar, Luis Roldão, and Dzmitry Tsishkou.
\newblock Planerf: Svd unsupervised 3d plane regularization for nerf large-scale urban scene reconstruction.
\newblock In \emph{3DV}, 2024{\natexlab{a}}.

\bibitem[Wang et~al.(2022)Wang, Wang, Long, Theobalt, Komura, Liu, and Wang]{wang2022neuris}
Jiepeng Wang, Peng Wang, Xiaoxiao Long, Christian Theobalt, Taku Komura, Lingjie Liu, and Wenping Wang.
\newblock Neuris: Neural reconstruction of indoor scenes using normal priors.
\newblock In \emph{Computer Vision--ECCV 2022: 17th European Conference, Tel Aviv, Israel, October 23--27, 2022, Proceedings, Part XXXII}, pages 139--155. Springer, 2022.

\bibitem[Wang et~al.(2021)Wang, Liu, Liu, Theobalt, Komura, and Wang]{NeuS}
Peng Wang, Lingjie Liu, Yuan Liu, Christian Theobalt, Taku Komura, and Wenping Wang.
\newblock Neus: Learning neural implicit surfaces by volume rendering for multi-view reconstruction.
\newblock In \emph{NeurIPS}, 2021.

\bibitem[Wang et~al.(2023{\natexlab{a}})Wang, Han, Habermann, Daniilidis, Theobalt, and Liu]{neus2}
Yiming Wang, Qin Han, Marc Habermann, Kostas Daniilidis, Christian Theobalt, and Lingjie Liu.
\newblock Neus2: Fast learning of neural implicit surfaces for multi-view reconstruction.
\newblock In \emph{ICCV}, 2023{\natexlab{a}}.

\bibitem[Wang et~al.(2024{\natexlab{b}})Wang, Huang, Ye, Zhang, Ouyang, and He]{wang2024neurodin}
Yifan Wang, Di Huang, Weicai Ye, Guofeng Zhang, Wanli Ouyang, and Tong He.
\newblock Neurodin: A two-stage framework for high-fidelity neural surface reconstruction.
\newblock \emph{arXiv preprint arXiv:2408.10178}, 2024{\natexlab{b}}.

\bibitem[Wang et~al.(2024{\natexlab{c}})Wang, Tan, Navab, and Tombari]{wang2024raneus}
Yida Wang, David~Joseph Tan, Nassir Navab, and Federico Tombari.
\newblock Raneus: Ray-adaptive neural surface reconstruction.
\newblock In \emph{2024 International Conference on 3D Vision (3DV)}, pages 53--63. IEEE, 2024{\natexlab{c}}.

\bibitem[Wang et~al.(2023{\natexlab{b}})Wang, Shen, Gao, Huang, Munkberg, Hasselgren, Gojcic, Chen, and Fidler]{fegr}
Zian Wang, Tianchang Shen, Jun Gao, Shengyu Huang, Jacob Munkberg, Jon Hasselgren, Zan Gojcic, Wenzheng Chen, and Sanja Fidler.
\newblock Neural fields meet explicit geometric representation for inverse rendering of urban scenes.
\newblock In \emph{CVPR}, 2023{\natexlab{b}}.

\bibitem[Wang et~al.(2023{\natexlab{c}})Wang, Shen, Nimier-David, Sharp, Gao, Keller, Fidler, Müller, and Gojcic]{wang2023adaptive}
Zian Wang, Tianchang Shen, Merlin Nimier-David, Nicholas Sharp, Jun Gao, Alexander Keller, Sanja Fidler, Thomas Müller, and Zan Gojcic.
\newblock Adaptive shells for efficient neural radiance field rendering.
\newblock In \emph{SIGGRAPH Asia}, 2023{\natexlab{c}}.

\bibitem[Xiao et~al.(2021)Xiao, Shao, Hao, Zhang, Chai, Jiao, Li, Wu, Sun, Jiang, et~al.]{pandaset}
Pengchuan Xiao, Zhenlei Shao, Steven Hao, Zishuo Zhang, Xiaolin Chai, Judy Jiao, Zesong Li, Jian Wu, Kai Sun, Kun Jiang, et~al.
\newblock Pandaset: Advanced sensor suite dataset for autonomous driving.
\newblock In \emph{ITSC}, 2021.

\bibitem[Yang et~al.(2023)Yang, Chen, Wang, Manivasagam, Ma, Yang, and Urtasun]{yang2023unisim}
Ze Yang, Yun Chen, Jingkang Wang, Sivabalan Manivasagam, Wei-Chiu Ma, Anqi~Joyce Yang, and Raquel Urtasun.
\newblock Unisim: A neural closed-loop sensor simulator.
\newblock In \emph{Proceedings of the IEEE/CVF Conference on Computer Vision and Pattern Recognition}, pages 1389--1399, 2023.

\bibitem[Yariv et~al.(2021)Yariv, Gu, Kasten, and Lipman]{VolSDF}
Lior Yariv, Jiatao Gu, Yoni Kasten, and Yaron Lipman.
\newblock Volume rendering of neural implicit surfaces.
\newblock In \emph{NeurIPS}, 2021.

\bibitem[Yu et~al.(2022)Yu, Peng, Niemeyer, Sattler, and Geiger]{Yu2022MonoSDF}
Zehao Yu, Songyou Peng, Michael Niemeyer, Torsten Sattler, and Andreas Geiger.
\newblock Monosdf: Exploring monocular geometric cues for neural implicit surface reconstruction.
\newblock \emph{NeurIPS}, 2022.

\bibitem[Yu et~al.(2024)Yu, Sattler, and Geiger]{Yu2024GOF}
Zehao Yu, Torsten Sattler, and Andreas Geiger.
\newblock Gaussian opacity fields: Efficient and compact surface reconstruction in unbounded scenes.
\newblock \emph{arXiv preprint arXiv:2404.10772}, 2024.

\bibitem[Zhang et~al.(2023)Zhang, Hu, Wu, Zhao, Li, Zou, and Fan]{zhang2023towards}
Yongqiang Zhang, Zhipeng Hu, Haoqian Wu, Minda Zhao, Lincheng Li, Zhengxia Zou, and Changjie Fan.
\newblock Towards unbiased volume rendering of neural implicit surfaces with geometry priors.
\newblock In \emph{Proceedings of the IEEE/CVF Conference on Computer Vision and Pattern Recognition}, pages 4359--4368, 2023.

\end{thebibliography}
}

\end{document}